\documentclass{article}

\usepackage[table,svgnames]{xcolor}
\usepackage{iclr2023_conference,times}

\newtheorem{assumption}{Assumption}

\newtheorem{remark}{Remark}

\usepackage{amsmath,amsfonts,bm}

\def\eqref#1{equation~\ref{#1}}

\def\1{\bm{1}}

\DeclareMathAlphabet{\mathsfit}{\encodingdefault}{\sfdefault}{m}{sl}
\SetMathAlphabet{\mathsfit}{bold}{\encodingdefault}{\sfdefault}{bx}{n}

\def\gG{{\mathcal{G}}}

\def\gL{{\mathcal{L}}}

\def\gT{{\mathcal{T}}}

\def\gV{{\mathcal{V}}}

\def\gX{{\mathcal{X}}}

\newcommand{\E}{\mathop{\mathbb{E}}}

\newcommand{\R}{\mathbb{R}}

\newcommand{\Var}{\mathrm{Var}}

\DeclareMathOperator*{\argmin}{arg\,min}
\DeclareMathOperator*{\argsup}{arg\,sup}
\DeclareMathOperator*{\arginf}{arg\,inf}
\DeclareMathOperator*{\subjectto}{subject\;to}

\usepackage[T3,T1]{fontenc}
\DeclareSymbolFont{tipa}{T3}{cmr}{m}{n}
\DeclareMathAccent{\invbreve}{\mathalpha}{tipa}{16}

\newcommand{\optdual}{\ensuremath{\invbreve{f}}}
\newcommand{\optprimal}{\ensuremath{\breve{T}}}
\newcommand{\optconj}{\ensuremath{\breve{x}}}

\usepackage[utf8]{inputenc}
\usepackage[T1]{fontenc}
\usepackage{textcomp}
\usepackage{framed}

\usepackage{listings}
\usepackage{graphicx}
\usepackage{booktabs}
\usepackage{diagbox}
\usepackage{wrapfig}
\usepackage{caption}

\definecolor{linkcolor}{RGB}{74, 102, 146}
\definecolor{coral}{HTML}{F2545B}
\usepackage[colorlinks=true,allcolors=linkcolor,pageanchor=true,plainpages=false,pdfpagelabels,bookmarks,bookmarksnumbered]{hyperref}

\newcommand{\cellhi}{\cellcolor{RoyalBlue!15}}
\newcommand{\abbrev}[1]{{\color{gray} *[#1]}}
\newcommand{\pair}[2]{$#1$ {\color{gray}\footnotesize $\pm#2$}}
\newcommand{\abbrevtab}[1]{\llap{\abbrev{#1}\hspace{2.5mm}}}

\usepackage{graphicx}
\usepackage{tikz}
\usepackage{hyperref}
\usepackage{url}
\usepackage[edges]{forest}

\usepackage{algpseudocode}
\algnewcommand{\LeftCommentX}[1]{\Statex \(\triangleright\) #1}
\algnewcommand{\LeftComment}[1]{\State \(\triangleright\) #1}

\usepackage{tikz}
\usetikzlibrary{arrows,backgrounds,bayesnet,calc,matrix}
\usetikzlibrary{arrows.meta}

\newcommand{\cblock}[3]{
  \hspace{-1.5mm}
  \begin{tikzpicture}[node/.style={square, minimum size=10mm, thick, line width=0pt}]
    \node[fill={rgb,255:red,#1;green,#2;blue,#3}] () [] {};
  \end{tikzpicture}%
}

\usepackage{algorithm, algpseudocode}
\DeclareMathOperator\supp{supp}

\usepackage[nameinlink]{cleveref}
\Crefname{equation}{Eq.}{Eqs.}
\Crefname{section}{Sect.}{Sects.}
\Crefname{appendix}{App.}{Apps.}
\Crefname{definition}{Def.}{Defs.}
\Crefname{proposition}{Prop.}{Props.}
\Crefname{assumption}{Assumption}{Assumptions}

\usepackage{xspace}
\newcommand{\eg}{e.g.\xspace}
\newcommand{\ie}{i.e.\xspace}

\usepackage{mathtools}
\newcommand{\defeq}{\vcentcolon=}

\iclrfinalcopy
\begin{document}

\title{On amortizing convex conjugates \\ for optimal transport}
\author{Brandon Amos \\ Meta AI}
\maketitle

\begin{abstract}
  This paper focuses on computing the convex conjugate (also
  known as the Legendre–Fenchel conjugate or c-transform) that appears in
  Euclidean Wasserstein-2 optimal transport.
  This conjugation is considered difficult to compute and in practice,
  methods are limited by not being able to
  exactly conjugate the dual potentials in continuous space.
  To overcome this, the computation of the conjugate can be
  approximated with amortized optimization, which
  learns a model to predict the conjugate.
  I show that combining amortized approximations to the
  conjugate with a solver for fine-tuning
  significantly improves the quality of transport
  maps learned for the Wasserstein-2 benchmark by
  \citet{korotin2021neural} and is able to model many 2-dimensional
  couplings and flows considered in the literature.
  All baselines, methods, and solvers are publicly available at
  \url{http://github.com/facebookresearch/w2ot}.
\end{abstract}

\section{Introduction}
Optimal transportation
\citep{villani2009optimal,ambrosio2003lecture,santambrogio2015optimal,peyre2019computational}
is a thriving area of research that provides a way of
connecting and transporting between probability measures.
While optimal transport between discrete measures
is well-understood, \eg with Sinkhorn distances
\citep{cuturi2013sinkhorn}, optimal transport between continuous
measures is an open research topic actively being investigated
\citep{genevay2016stochastic,seguy2017large,taghvaei2019wasserstein,korotin2019wasserstein,makkuva2020optimal,fan2021scalable,asadulaev2022neural}.
Continuous OT has applications in
generative modeling \citep{arjovsky2017wasserstein,petzka2017regularization,wu2018wasserstein,liu2019wasserstein,cao2019multi,leygonie2019adversarial},
domain adaptation \citep{luo2018wgan,shen2018wasserstein,xie2019scalable},
barycenter computation \citep{li2020continuous,fan2020scalable,korotin2021continuous},
and biology \citep{bunne2021learning,bunne2022supervised,lubeck2022neural}.

\textbf{This paper focuses on estimating the Wasserstein-2 transport map}
between measures $\alpha$ and $\beta$ in
\emph{Euclidean} space, \ie $\supp(\alpha)=\supp(\beta)=\R^n$ with the
Euclidean distance as the transport cost.
The \emph{Wasserstein-2 transport map}, $\optprimal: \R^n\rightarrow \R^n$,
is the solution to \emph{Monge's primal formulation}:
\begin{equation}
  \optprimal \in \arginf_{T\in\gT(\alpha,\beta)} \E_{x\sim\alpha} \|x-T(x)\|^2_2,
  \label{eq:monge-primal}
\end{equation}
where $\gT(\alpha,\beta)\defeq \{T: T_\#\alpha=\beta\}$ is the set
of admissible couplings
and the \emph{push-forward operator} $\#$ is defined by
$T_{\#}\alpha(B) \defeq \alpha(T^{-1}(B))$
for a measure $\alpha$, measurable map $T$, and all measurable sets $B$.
$\optprimal$ exists and is unique under
general settings, \eg as in \citet[Theorem~1.17]{santambrogio2015optimal}, and is often difficult to
solve because of the coupling constraints $\gT$.
\textbf{Almost every computational method instead
  solves the Kantorovich dual},
\eg as formulated in
\citet[\S5]{villani2009optimal} and \citet[\S2.5]{peyre2019computational}.
This paper focuses on the dual associated with the
negative inner product cost
\citep[eq.~5.12]{villani2009optimal},
which introduces a
\emph{dual potential function} $f: \R^n\rightarrow\R$
and solves:
\looseness=-1
\begin{equation}
  \optdual\in\argsup_{f\in L^1(\alpha)}\; -\E_{x\sim\alpha}[f(x)] - \E_{y\sim\beta}[f^\star(y)]
  \label{eq:kantorovich-dual}
\end{equation}
where
$L^1(\alpha)$ is the space of measurable functions that
are Lebesgue-integrable over $\alpha$ and
$f^\star$ is the \emph{convex conjugate},
or \emph{Legendre-Fenchel} transform, of a function
$f$ defined by:
\begin{equation}
  f^\star(y)\defeq -\inf_{x\in\gX} J_f(x; y) \quad \text{with objective}\quad J_f(x; y)\defeq f(x)-\langle x, y\rangle.
  \label{eq:conj}
\end{equation}
$\optconj(y)$ denotes an optimal solution to \cref{eq:conj}.
Even though the \cref{eq:kantorovich-dual} searches over
functions in $L^1(\alpha)$, the \emph{optimal dual potential}
$\optdual$ is convex \citep[theorem~5.10]{villani2009optimal}.
When one of the measures has a density,
\citet[theorem~3.1]{brenier1991polar} and \citet{mccann1995existence}
relate
$\optdual$ to an optimal transport map $\optprimal$ for the primal
problem in \cref{eq:monge-primal} with
$\optprimal(x) = \nabla_x \optdual(x)$,
and the inverse of the transport map is given by
$\optprimal^{-1}(y)=\nabla_y \optdual^\star(y)$.

\textbf{Several foundational papers have proposed methods to
approximate the dual potential $f$ with a neural network
and learn it by optimizing \cref{eq:kantorovich-dual}:}
\citet{taghvaei2019wasserstein,korotin2019wasserstein,makkuva2020optimal}
parameterize $f$ as an \emph{input-convex neural network}
\citep{amos2017input}, which can universally
represent any convex function with enough capacity \citep{huang2020convex}.
Other works explore parameterizing $f$ as a \emph{non-convex}
neural network \citep{nhan2019threeplayer,korotin2021neural,rout2021generative}.

\textbf{Efficiently solving the conjugation operation
  in \cref{eq:conj} is the key computational challenge
  to solving the Kantorovich dual in \cref{eq:kantorovich-dual}}
and is an important design choice.
Exactly computing the conjugate as done in
\citet{taghvaei2019wasserstein} is considered computationally
challenging and approximating it as in
\citet{korotin2019wasserstein,makkuva2020optimal,nhan2019threeplayer,korotin2021neural,rout2021generative}
may be unstable.
\citet{korotin2021neural} fortifies this observation:

\begin{leftbar}
\emph{The [exact conjugate] solver is slow since each optimization
step solves a hard subproblem for computing [the conjugate].
[Solvers that approximate the conjugate] are also hard to optimize:
they either diverge from the start or diverge after converging to a
nearly-optimal saddle point.}
\end{leftbar}

\textbf{In contrast to these statements on the difficulty of exactly
estimating the conjugate operation, I will show in this paper
that computing the (near-)exact conjugate is easy.}
My key insight is that the approximate, \ie \emph{amortized},
conjugation methods can be combined with a fine-tuning
procedure using the approximate solution as a starting point.
\Cref{sec:amortization} discusses the amortization design choices
and \cref{sec:amor:cycle} presents a new amortization perspective on
the cycle consistency term used in Wasserstein-2 generative
networks \citep{korotin2019wasserstein}, which was previously not seen in this way.
\textbf{\Cref{sec:exp:w2} shows that amortizing and fine-tuning
the conjugate results in state-of-the-art
performance in \emph{all} of the tasks proposed
in the Wasserstein-2 benchmark by \citet{korotin2021neural}.}
Amortization with fine-tuning also nicely models synthetic settings (\cref{sec:demos}),
including for learning a single-block potential flow \emph{without} using the likelihood.

\section{Learning dual potentials: a conjugation perspective}
\label{sec:learning-potentials}

This section reviews the standard methods of learning
parameterized dual potentials to solve
\cref{eq:kantorovich-dual}.
The first step is to re-cast the Kantorovich dual
problem \cref{eq:kantorovich-dual}
as being over a \emph{parametric} family of potentials
$f_\theta$ with parameter $\theta$ as an
\emph{input-convex neural network} \citep{amos2017input}
or a more general non-convex neural network.
\citet{taghvaei2019wasserstein,makkuva2020optimal}
have laid the foundations for optimizing the parametric
potentials for the dual objective with
\begin{equation}
  \max_\theta \gV(\theta)
  \quad \text{where}\quad \gV(\theta)\defeq -\E_{x\sim\alpha}[f_\theta(x)] - \E_{y\sim\beta}[f_\theta^\star(y)] =
  -\E_{x\sim\alpha}[f_\theta(x)] + \E_{y\sim\beta}\left[J_{f_\theta}(\optconj(y))\right],
  \label{eq:kantorovich-dual-parametric}
\end{equation}
where
$J$ is the objective to the conjugate optimization problem in \cref{eq:conj},
$\optconj(y)$ is the solution to the convex conjugate,
and \cref{eq:kantorovich-dual-parametric} assumes a finite solution
to \cref{eq:kantorovich-dual} exists and replaces the $\sup$ with
a $\max$.
\citet{taghvaei2019wasserstein} show that the model can be \emph{learned},
\ie the optimal parameters can be found,
by taking gradient steps of the dual with respect
to the parameters of the potential, \ie using $\nabla_\theta\gV$.
This derivative going through the loss and conjugation operation
can be obtained by applying \emph{Danskin's envelope theorem}
\citep{danskin1966theory,bertsekas1971control} and
results in only needing derivatives of the potential:
\begin{equation}
  \begin{aligned}
  \nabla_\theta \gV(\theta) &= \nabla_\theta\left[-\E_{x\sim\alpha}[f_\theta(x)] +
    \E_{y\sim\beta}\left[J_{f_\theta}(\optconj(y))\right]\right] \\
  & = -\E_{x\sim\alpha}[\nabla_\theta f_\theta(x)] +
    \E_{y\sim\beta} [\nabla_\theta f_\theta(\optconj(y)) ]
  \label{eq:kantorovich-dual-derivative}
  \end{aligned}
\end{equation}
where $\optconj(y)$ is not differentiated through.

\begin{assumption}
  A standard assumption is that the conjugate is smooth with a
  well-defined $\argmin$.
  This has been shown to hold when $f$ is strongly convex,
  \eg in \citet{kakade2009duality},
  or when $f$ is
  essentially strictly convex \citep[theorem~26.3]{rockafellar2015convex}.
  \label{conj-unique}
\end{assumption}

In practice, \cref{conj-unique} is not guaranteed, \eg
non-convex potentials may have a parameterization that results
in the conjugate taking infinite values in regions.
The dual objective in \cref{eq:kantorovich-dual} and
\cref{eq:kantorovich-dual-parametric} discourage the conjugate
from diverging as the supremum involves the negation of the conjugate.

\begin{remark}
  In \cref{eq:kantorovich-dual-parametric}, the dual potential $f$ associated
  with the $\alpha$ measure's constraints is the central object that is
  parameterized and learned, and the dual potential associated with
  the $\beta$ measure is given by the conjugate $f^\star$ and does
  not require separately learning.
  Because of the symmetry of
  \cref{eq:monge-primal}, the order can also be
  \textbf{reversed} as in \citet{korotin2021continuous} so
  that the duals associated with the $\beta$ measure are the ones
  directly parameterized, but we will not consider doing this.
  Potentials associated with both measures can also be
  parameterized and we will next see that it is the most natural
  to think about the model associated with the conjugate as an
  amortization model.
\end{remark}

\begin{remark}
  The dual objective $\gV$ can be upper-bounded by replacing
  $\optconj$ with any approximation because any
  sub-optimal solution to the conjugation objective provides
  an upper-bound to the true objective, \ie
  $J(\optconj(y); y) \leq J(x; y)$ for all $x$.
  In practice, maximizing a loose upper-bound can cause
  significant divergence issues as the potential
  can start over-optimizing the objective.
  \label{rk:bound}
\end{remark}

Computing the updates to the dual potential's parameters in
\cref{eq:kantorovich-dual-derivative} is a well-defined machine
learning setup given a parameterization of the potential $f_\theta$,
but is often computationally bottlenecked by the conjugate operation.
Because of this bottleneck, many existing works resort to
\emph{amortizing the conjugate} by predicting the solution with a model
$\tilde x_\phi(y)$.
I overview the design choices behind amortizing the conjugate
in \cref{sec:amortization}, and then go on in
\cref{sec:numerical-solvers} to show that it is reasonable
to fine-tune the amortized predictions with an
\emph{explicit solver}
$\textsc{conjugate}(f, y, x_{\rm init}=\tilde x_\phi(y))$.
\Cref{alg:w2-alg} summarizes how to learn a dual potential
with an amortized and fine-tuned conjugate.

\section{Amortizing convex conjugates: modeling and losses}
\label{sec:amortization}

This section scopes to \emph{predicting} an approximate
solution to the conjugate optimization problem in \cref{eq:conj}.
This is an instance of \emph{amortized optimization} methods
which predict the solution to a family of
optimization problems that are repeatedly solved
\citep{shu2017amortized,chen2021learning,amos2022tutorial}.
Amortization is sensible here because
the conjugate is repeatedly solved for $y\sim\beta$
every time the dual $\gV$ from
\cref{eq:kantorovich-dual-parametric} is evaluated across a batch.
Using the basic setup from \citet{amos2022tutorial},
I call a prediction to the solution of \cref{eq:conj}
the \emph{amortization model} $\tilde x_\varphi(y)$,
which is parameterized by some $\varphi$.
The goal is to make the amortization model's prediction
match the true conjugate solution,
\ie $\tilde x_\phi(y)\approx\optconj(y)$,
for samples $y\sim\beta$.
In other words, amortization uses a model to simultaneously
solve \emph{all} of the conjugate optimization problems.
There are two main design choices:
\cref{sec:amor:models} discusses \emph{parameterizing the
amortization model} and
\cref{sec:amor:losses}
overviews \emph{amortization losses}.

\subsection{Parameterizing a conjugate amortization model}
\label{sec:amor:models}
The \emph{amortization model} $\tilde x_\varphi(y)$ maps
a point $y\sim\beta$ to a solution to the conjugate
in \cref{eq:conj}, \ie $\tilde x_\varphi: \R^n\rightarrow \R^n$
and the goal is for $\tilde x_\varphi(y) \approx \optconj(y)$.
In this paper, I take standard potential models
further described in \cref{app:models}
and keep them fixed to ablate across the
amortization loss and fine-tuning choices.
The main categories are:

\begin{enumerate}
\item $\tilde x_\varphi: \R^n\rightarrow \R^n$
  \emph{directly maps to the solution} of
  \cref{eq:conj} with a multilayer perceptron (MLP) as in \citet{nhan2019threeplayer},
  or a U-Net \citep{ronneberger2015u} for image-based transport.
  These are also used in parts of \citet{korotin2021neural}.
  \item $\tilde x_\varphi = \nabla_y g_\varphi$
    is the \emph{gradient of a function} $g_\varphi: \R^n\rightarrow\R$.
    \citet{korotin2019wasserstein,makkuva2020optimal} parameterize
    $g_\varphi$ as an input-convex neural network, and some methods of
    \citet{korotin2021neural} parametrize $g_\varphi$ as a ResNet \citep{he2016identity}.
    This is well-motivated because the $\argmin$ of a convex conjugate
    is the derivative, \ie
    $\optconj(y)=\nabla_y f^\star(y)$.
\end{enumerate}

\begin{algorithm}[t]
  \caption{Learning Wasserstein-2 dual potentials with amortized and fine-tuned conjugation}
  \begin{algorithmic}
    \State \textbf{Inputs:} Measures $\alpha$ and $\beta$ to couple,
    initial dual potential $f_\theta$, and initial amortization model $\tilde x_\varphi$
    \While {unconverged}
    \State \emph{Sample batches} $\{x_j\}\sim\alpha$ and $\{y_j\}\sim\beta$ indexed by $j\in[N]$
    \State Obtain the \emph{amortized prediction} of the conjugate $\tilde x_\varphi(y_j)$
    \State \emph{Fine-tune the prediction} by numerically solving $\optconj(y_j)=\textsc{conjugate}(f, y_j, x_{\rm init}=\tilde x_\varphi(y_j))$
    \State \emph{Update the potential} with a gradient estimate of the dual in
        \cref{eq:kantorovich-dual-derivative}, \ie $\nabla_\theta \gV$
        \State \emph{Update the amortization model} with a gradient estimate
        of a loss from \cref{sec:amortization}, \ie $\nabla_\varphi \gL$
    \EndWhile
    \State\Return optimal dual potentials $f_\theta$ and conjugate amortization model $\tilde x_\varphi$
  \end{algorithmic}
  \label{alg:w2-alg}
\end{algorithm}

\newpage
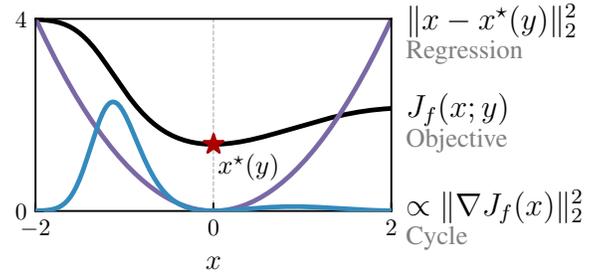
\begin{wrapfigure}{r}{2.2in}
  \vspace{-12mm}
  \centering
  \begin{tikzpicture}[every node/.style={align=left,anchor=south west}]
    \node (im) {\includegraphics[width=2.0in]{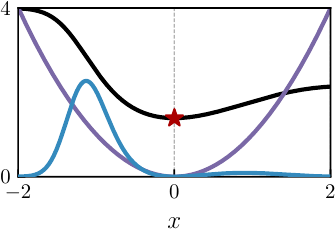}};
    \node at (52mm, 2.8mm) () {\large $\propto\|\nabla J_f(x)\|_2^2$ \\ {\color{gray}Cycle}};
    \node at (52mm, 15.8mm) () {\large $J_f(x; y)$ \\ {\color{gray}Objective}};
    \node at (52mm, 27.5mm) () {\large $\|x-x^\star(y)\|_2^2$ \\ {\color{gray}Regression}};
    \node at (27mm,12mm) () {$x^\star(y)$};
  \end{tikzpicture}%
  \vspace{-3mm}
  \captionsetup{font={small}}
  \caption{Conjugate amortization losses.}
  \label{fig:amor-choices}
\end{wrapfigure}
\subsection{Conjugate amortization loss choices}
\label{sec:amor:losses}
We now turn to the design choice of what loss to optimize
so that the conjugate amortization model $\tilde x_\varphi$ best-predicts
the solution to the conjugate. In all cases, the loss is
differentiable and $\varphi$ is optimized with a gradient-based optimizer.
I present an amortization perspective of methods not previously
presented as amortization methods, which is useful to
help think about improving the amortized predictions
with the fine-tuning and exact solvers in \cref{sec:numerical-solvers}.
\Cref{fig:amor-choices} illustrates the main loss choices.

\subsubsection{Objective-based amortization}
\label{sec:amor:obj}
\citet{nhan2019threeplayer} propose to make the amortized
prediction optimal on the
\emph{conjugation objective} $J_f$ from \cref{eq:conj}
across samples from $\beta$, \ie:
\begin{equation}
  \min_\varphi \gL_{\rm obj}(\varphi) \; \text{where}\; \gL_{\rm obj}(\varphi)\defeq \E_{y\sim \beta} J_f(\tilde x_\varphi(y); y).
  \label{eq:amor-obj-opt}
\end{equation}
We refer to $\gL_{\rm obj}$ as \emph{objective-based} amortization
and solve \cref{eq:amor-obj-opt} by taking gradient
steps $\nabla_\varphi \gL_{\rm obj}$ using a Monte-Carlo estimate
of the expectation.

\begin{remark}
  \label{rk:makkuva-amortization}
  The maximin method proposed in \citet[theorem~3.3]{makkuva2020optimal} is
  equivalent to maximizing an upper-bound to the dual loss $\gV$ with respect
  to $\theta$ of a potential $f_\theta$ and minimizing the objective-based
  amortization loss $\gL_{\rm obj}$ with respect to $\varphi$ of an
  amortization model $\tilde x_\varphi\defeq\nabla g_\varphi$.
  Their formulation replaces the exact conjugate $\optconj$ in
  \cref{eq:kantorovich-dual-parametric}
  with an approximation $\tilde x_\varphi$, \ie:
  \begin{equation}
    \max_\theta\min_\varphi \gV_{\rm MM}(\theta, \varphi) \; {\rm where} \;
    \gV_{\rm MM}(\theta, \varphi) \defeq -\E_{x\sim\alpha}[f_\theta(x)] + \E_{y\sim\beta}[J_{f_\theta}(\tilde x_\varphi(y); y)].
    \label{eq:makkuva-motivation}
  \end{equation}
  \citet{makkuva2020optimal} propose to optimize $\gV_{\rm MM}$ with gradient ascent-descent steps.
  For optimizing $\theta$, $\gV_{\rm MM}(\theta, \varphi)$ is an upper bound
  on the true dual objective $\gV(\theta)$ as discussed in \cref{rk:bound}
  with equality if and only if $\tilde x_\varphi=\optconj$.
  \textbf{Evaluating the inner optimization step is exactly
    the objective-based amortization update,} \ie,
  $\nabla_\varphi \gV_{\rm MM}(\theta, \varphi) = \nabla_\varphi \gL_{\rm obj}(\varphi) = \nabla_\varphi J_{f_\theta}(\tilde x_\varphi(y); y)$.
\end{remark}

\begin{remark}
  Suboptimal predictions of the conjugate often leads to
  a divergent upper bound on $\gV(\theta)$.
  \citet[algorithm~1]{makkuva2020optimal} propose to fix this
  by running more updates on the amortization model.
  In \cref{sec:numerical-solvers},
  I propose fine-tuning as an alternative to
  obtain near-exact conjugates.
\end{remark}

\subsubsection{First-order optimality amortization: cycle consistency and W2GN}
\label{sec:amor:cycle}

An alternative to optimizing the dual objective directly as in
\cref{eq:amor-obj-opt} is to
optimize for the \emph{first-order optimality condition}.
\Cref{eq:conj} is an unconstrained minimization problem, so the
first-order optimality condition is that the derivative of the objective is zero,
\ie $\nabla_x J_f(x; y) = \nabla_x f(x) - y = 0$.
The conjugate amortization model can be optimized for the \emph{residual norm} of this condition with
\begin{equation}
  \min_\varphi \gL_{\rm cycle}(\varphi) \; \text{where}\; \gL_{\rm cycle}(\varphi)\defeq \E_{y\sim \beta} \|\nabla_x J_f(\tilde x_\varphi(y); y)\|_2^2 = \E_{y\sim\beta} \|\nabla_x f(\tilde x_\varphi(y)) - y\|_2^2.
  \label{eq:amor-obj-cycle}
\end{equation}

\begin{remark}
W2GN \citep{korotin2019wasserstein} is equivalent to maximizing an upper-bound to
the dual loss $\gV$ with respect to $\theta$ of a potential $f_\theta$ and minimizing the
first-order amortization loss $\gL_{\rm cycle}$ with respect to $\varphi$ of an
conjugate amortization model $\tilde x_\varphi\defeq\nabla g_\varphi$.
\citet{korotin2019wasserstein} originally motivated the cycle
consistency term from its use in cross-domain generative modeling
\citet{zhu2017unpaired} and \textbf{\cref{eq:amor-obj-cycle} shows an alternative
way of deriving the cycle consistency term by amortizing the
first-order optimality conditions of the conjugate.}
\label{rk:w2gn-amortization}
\end{remark}

\begin{remark}
  The formulation in \citet{korotin2019wasserstein} does
  \textbf{not} disconnect $f_\theta$ when optimizing
  the cycle loss in \cref{eq:amor-obj-cycle}.
  From an amortization perspective, this performs amortization
  by updating $f_\theta$ to have a solution closer to $\tilde x_\varphi$
  rather than the usual amortization setting of updating
  $\tilde x_\varphi$ to make a prediction closer to the solution of $f_\theta$.
  In my experiments, updating $f_\theta$ with the amortization
  term seems to help when not fine-tuning the conjugate
  to be exact, but not when using the exact conjugates.
  \looseness=-1
\end{remark}

\begin{remark}
  \citet{korotin2019wasserstein} and followup papers such as \citet{korotin2021continuous}
  state that they do not perform maximin optimization as in
  \cref{eq:makkuva-motivation}
  from \citet{makkuva2020optimal} because they replace the inner optimization
  of the conjugate with an approximation.
  \textbf{I disagree that the main distinction between these methods
    should be based on their formulation as a maximin optimization problem.}
  I instead propose that the main difference between their losses
  is how they amortize the convex conjugate:
  \citet{makkuva2020optimal} use the objective-based loss in
  \cref{eq:amor-obj-opt} while
  \citet{korotin2019wasserstein} use the first-order optimality
  condition (\cref{eq:amor-obj-cycle}).
  \Cref{sec:exp:w2} shows that adding fine-tuning and exact conjugates
  to both of these methods makes their performance match in most
  cases.
  \looseness=-1
\end{remark}

\begin{remark}
  Optimizing for the first-order optimality conditions may not be ideal
  for non-convex conjugate objectives as inflection points with
  a near-zero derivative may not be a global minimum of \cref{eq:conj}.
  The left and right regions of \cref{fig:amor-choices} illustrate this.
\end{remark}

\subsubsection{Regression-based amortization}
\label{sec:amor:reg}

The previous objective and first-order amortization methods
locally refine the model's prediction using local derivative information.
The conjugate amortization model can also be trained by regressing onto ground-truth solutions
when they are available, \ie
\begin{equation}
  \min_\varphi \gL_{\rm reg}(\varphi) \; \text{where}\; \gL_{\rm reg}(\varphi)\defeq \E_{y\sim \beta} \|\tilde x_\varphi(y)-\optconj(y)\|_2^2
  \label{eq:amor-obj-reg}.
\end{equation}
This regression loss is the most useful when approximations
to the conjugate are computationally easy to obtain,
\eg with a method described in \cref{sec:numerical-solvers}.
$\gL_{\rm reg}$ gives the amortization model information
about where the globally optimal solution is rather
than requiring it to only locally search over the conjugate's
objective $J$.

\section{Numerical solvers for exact conjugates and fine tuning}
\label{sec:numerical-solvers}
\begin{wrapfigure}{r}{2in}
\vspace{-11mm}
\begin{minipage}{\linewidth}
\begin{algorithm}[H]
\captionsetup{font={small}}
\caption{$\textsc{conjugate}(f, y, x_{\rm init})$}
\label{alg:conjugate}
\begin{algorithmic}
\footnotesize
\State $x\leftarrow x_{\rm init}$
\While {unconverged}
\State Update $x$ with $\nabla_x J_f(x; y)$
\EndWhile
\State\Return optimal $\optconj(y)=x$
\end{algorithmic}
\end{algorithm}
\end{minipage}
\vspace{-8mm}
\end{wrapfigure}

In the Euclidean Wasserstein-2 setting, the conjugation operation in \cref{eq:conj}
is a continuous and unconstrained optimization problem over
a possibly non-convex potential $f$.
It is usually implemented with a method using first-order information for the
update in \cref{alg:conjugate}, such as:
\vspace{3mm}

\begin{enumerate}
\item \emph{Adam} \citep{kingma2014Adam} is an adaptive
  first-order optimizer for high-dimensional optimization problems
  and is used for the exact conjugations in \citet{korotin2021neural}.
  \textbf{Note:} Adam here is for
  \cref{alg:conjugate} and is \emph{not} performing parameter optimization.
\item \emph{L-BFGS} \citep{liu1989limited} is a quasi-Newton method
  for optimizing unconstrained convex functions.
  \Cref{sec:lbfgs} discusses more implementation details behind
  setting up L-BFGS efficiently to run on the batches of optimization
  problems considered here. Choosing the line search method is the
  most crucial part as the conditional nature of some line searches may
  be prohibitive over batches.
  \Cref{tab:lbfgs-line-searches} shows that an Armijo search often works
  well to obtain approximate solutions.
\end{enumerate}

\section{Experimental results on the Wasserstein-2 benchmark}
\label{sec:exp:w2}
I have focused most of the experimental investigations on the Wasserstein-2 benchmark
\citep{korotin2021neural} because it provides a concrete evaluation
setting with established baselines for learning potentials for
Euclidean Wasserstein-2 optimal transport.
The tasks in the benchmark have known (ground-truth) optimal transport
maps and include transporting between:
1) high-dimensional (HD) mixtures of Gaussians, and
2) samples from generative models trained
on CelebA \citep{liu2015deep}.
The main evaluation metric is
the \emph{unexplained variance percentage} ($\gL^2$-UVP) metric from
\citep{korotin2019wasserstein}, which compares a candidate map
$T$ to the ground truth map $T^\star$ with:
\begin{equation}
  \label{eq:l2-uvp}
  \gL^2\text{-UVP}(T; \alpha, \beta)\defeq100\cdot\|T-T^\star\|^2_{\gL^2(\alpha)}/\Var(\beta)\%.
\end{equation}
In all of the experimental results, I report the \emph{final} $\gL^2$-UVP evaluated
with 16384 samples at the end of training, and average the results
over 10 trials.
\Cref{app:w2} further details the experimental setup.
My original motivation for running these experiments was to understand
how ablating the amortization losses and fine-tuning options impacts
the final $\gL^2$-UVP performance of the learned potential.

\begin{figure}[b]
  \centering
  \includegraphics[width=\textwidth]{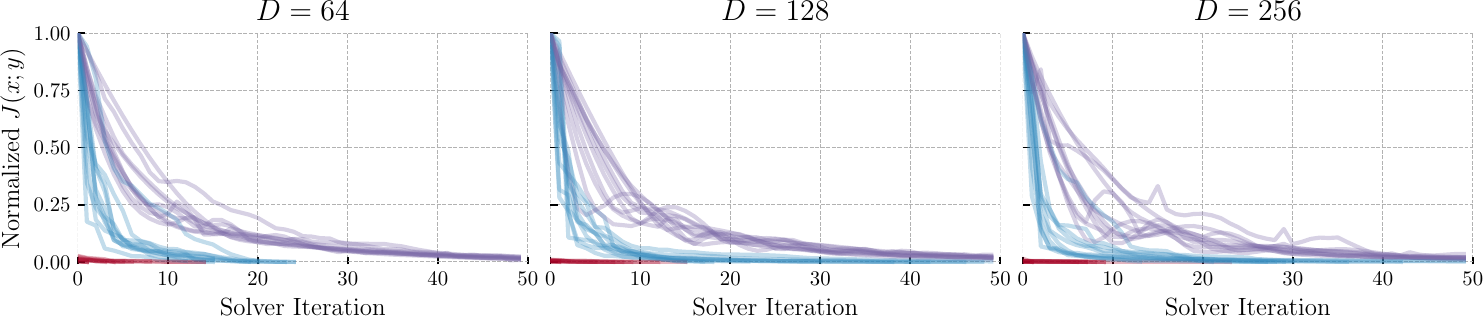}
  \cblock{166}{6}{40} Amortized Initialization + L-BFGS \hspace{2mm}
  \cblock{52}{138}{189} L-BFGS \hspace{2mm}
  \cblock{128}{114}{179} Adam
  \caption{Conjugate solver convergence on the HD benchmarks
    with an ICNN potential.}
  \label{fig:conj-convergence-icnn-hd}
  \vspace{-6mm}
\end{figure}

\textbf{The main experimental takeaway of this paper is that
fine-tuning the amortized conjugate with a solver
significantly improves the learned transport maps.}
\Cref{tab:hd,tab:celeba} report that amortizing and fine-tuning
the conjugate improves the $\gL^2$-UVP performance by a factor of
1.8 to 4.4 over the previously best-known results on the benchmark.
\Cref{app:w2:runtimes} shows that
the conjugate can often be fine-tuned within 100ms per batch
of 1024 examples on an NVIDIA Tesla V100 GPU,
\cref{fig:conj-convergence-icnn-hd,app:convergence} compare
Adam and L-BFGS for solving the conjugation.
The following remarks further summarize the results from these experiments:

\begin{remark}
  \textbf{With fine-tuning, the choice of regression or
    objective-based amortization doesn't significantly impact
    the $\gL^2$-UVP of the final potential.}
  This is because fine-tuning is usually able to find the
  optimal conjugates from the predicted starting points.
\end{remark}

\begin{remark}
  My re-implementation of W2GN \citep{korotin2019wasserstein},
  which uses cycle consistency amortization with no fine-tuning,
  often outperforms the results reported in \citet{korotin2021neural}.
  This is likely due to differences in the base potential and
  conjugate amortization models.
\end{remark}

\begin{remark}
  Cycle consistency sometimes provides difficult starting points for
  the fine-tuning methods, especially for L-BFGS.
  When learning non-convex potentials, this poor performance is
  likely related to the fact that \textbf{Newton methods are known to
  be difficult for saddle points} \citep{dauphin2014identifying}.
  Combining cycle consistency, which tries to find
  a point where the derivative is zero, with L-BFGS, which
  also tries to find a point where the derivative is zero,
  results in finding suboptimal inflection points of the potential
  rather than the true minimizer.
\end{remark}

\begin{remark}
  The performance of the methods using objective-based
  amortization without fine-tuning, as done in \citet{taghvaei2019wasserstein},
  is lower than the performance reported in \citet{korotin2021neural}.
  This is because I do not run multiple inner updates to
  update the conjugate amortization model.
  I instead advocate for fine-tuning the conjugate predictions
  with a known solver, eliminating the need for a hyper-parameter of the number
  of inner iterations that needs to be delicately tuned to make
  sure the amortized prediction alone does not diverge too much
  from the true conjugate.
\end{remark}

\begin{table}[t]
  \vspace{-5mm}
\caption{Comparison of $\gL^2$-UVP on the high-dimensional tasks from
  the Wasserstein-2 benchmark by \citet{korotin2021neural}, where
  \abbrev{the gray tags} denote their results.
  I report the mean and standard deviation across 10 trials.
  \textbf{Fine-tuning the amortized prediction with L-BFGS
  or Adam consistently improves the quality of the learned potential.}
}
\label{tab:hd}
\resizebox{1.\linewidth}{!}{
  \begin{tabular}{r@{}cc|cccccccc}
    \multicolumn{11}{l}{\textbf{Baselines} from \citet{korotin2021neural}} \\ \toprule
    & Amortization loss & Conjugate solver & $n=2$ & $n=4$ & $n=8$ & $n=16$ & $n=32$ & $n=64$ & $n=128$ & $n=256$ \\\midrule
\abbrevtab{W2} & Cycle & {\color{gray}{None}} & 0.1 &    0.7 &    2.6 &    3.3 &    6.0 &    7.2 &    2.0 &    2.7 \\
\abbrevtab{MMv1} & {\color{gray}{None}} & Adam &    0.2 &    1.0 &    1.8 &    1.4 &    6.9 &    8.1 &    2.2 &    2.6 \\
\abbrevtab{MMv2} & Objective & {\color{gray}{None}} &   0.1 &    0.68 &    2.2 &    3.1 &    5.3 &   10.1&    3.2 &    2.7 \\
\abbrevtab{MM} & Objective & {\color{gray}None} &    0.1 &    0.3 &    0.9 &    2.2 &    4.2 &    3.2 &    3.1 &    4.1 \\
\bottomrule \\[2mm]

    \multicolumn{11}{l}{\textbf{Potential model:} the input convex neural network described in \cref{app:icnn}} \textbf{Amortization model:} the MLP described in \cref{app:init_nn} \\
\toprule
& Amortization loss & Conjugate solver & $n=2$ & $n=4$ & $n=8$ & $n=16$ & $n=32$ & $n=64$ & $n=128$ & $n=256$ \\\midrule
& Cycle & {\color{gray}None}     &  \pair{0.28}{0.09} &  \pair{0.90}{0.11} &  \pair{2.23}{0.20} &  \pair{3.03}{0.06} &  \pair{5.32}{0.14} &  \pair{8.79}{0.16} &  \pair{5.66}{0.45} &  \pair{4.34}{0.14} \\
& Objective & {\color{gray}None} &  \pair{0.27}{0.09} &  \pair{0.78}{0.12} &  \pair{1.78}{0.26} &  \pair{2.00}{0.11} &               {\color{gray}>100} &               {\color{gray}>100} &               {\color{gray}>100} &               {\color{gray}>100} \\ \midrule
& Cycle & L-BFGS                 &  \pair{0.26}{0.09} &  \pair{0.77}{0.11} &  \pair{1.63}{0.28} &  \pair{1.15}{0.14} &  \pair{2.02}{0.10} &  \pair{4.48}{0.89} &  \pair{1.65}{0.10} &  \pair{5.93}{9.43} \\
& Objective & L-BFGS             &  \pair{0.26}{0.09} &  \pair{0.79}{0.12} &  \pair{1.63}{0.30} &  \pair{1.12}{0.11} &  \pair{1.92}{0.19} &  \pair{4.40}{0.79} &  \pair{1.64}{0.11} &  \pair{2.24}{0.13} \\
& Regression & L-BFGS            &  \pair{0.26}{0.09} &  \pair{0.78}{0.12} &  \pair{1.64}{0.29} &  \pair{1.14}{0.12} &  \pair{1.93}{0.20} &  \pair{4.41}{0.74} &  \pair{1.69}{0.11} &  \pair{2.21}{0.15} \\ \midrule
& Cycle & Adam      &  \pair{0.26}{0.09} &  \pair{0.79}{0.11} &  \pair{1.62}{0.29} &  \pair{1.14}{0.12} &  \pair{1.95}{0.21} &  \pair{4.55}{0.62} &  \pair{1.88}{0.26} &               {\color{gray}>100} \\
& Objective & Adam  &  \pair{0.26}{0.09} &  \pair{0.79}{0.14} &  \pair{1.62}{0.31} &  \pair{1.08}{0.14} &  \pair{1.89}{0.19} &  \pair{4.23}{0.76} &  \pair{1.59}{0.12} &  \pair{1.99}{0.15} \\
& Regression & Adam &  \pair{0.35}{0.07} &  \pair{0.81}{0.12} &  \pair{1.61}{0.32} &  \pair{1.09}{0.11} &  \pair{1.85}{0.20} &  \pair{4.42}{0.68} &  \pair{1.63}{0.08} &  \pair{1.99}{0.16} \\
\bottomrule \\[2mm]

\multicolumn{11}{l}{\textbf{Potential model:} the non-convex neural network (MLP) described in \cref{app:potential_nn}} \textbf{Amortization model:} the MLP described in \cref{app:init_nn} \\
\toprule
& Amortization loss & Conjugate solver & $n=2$ & $n=4$ & $n=8$ & $n=16$ & $n=32$ & $n=64$ & $n=128$ & $n=256$ \\\midrule
& Cycle & {\color{gray}None}     &  \pair{0.05}{0.00} &  \pair{0.35}{0.01} &  \pair{1.51}{0.08} &               {\color{gray}>100} &               {\color{gray}>100} &               {\color{gray}>100} &               {\color{gray}>100} &               {\color{gray}>100} \\
& Objective & {\color{gray}None} &               {\color{gray}>100} &               {\color{gray}>100} &               {\color{gray}>100} &               {\color{gray}>100} &               {\color{gray}>100} &               {\color{gray}>100} &               {\color{gray}>100} &               {\color{gray}>100} \\ \midrule
& Cycle & L-BFGS                 &               {\color{gray}>100} &               {\color{gray}>100} &               {\color{gray}>100} &               {\color{gray}>100} &               {\color{gray}>100} &               {\color{gray}>100} &               {\color{gray}>100} &               {\color{gray}>100} \\
& Objective & L-BFGS             &  \cellhi \pair{0.03}{0.00} &  \cellhi \pair{0.22}{0.01} &  \cellhi \pair{0.60}{0.03} &  \cellhi \pair{0.80}{0.11} &  \cellhi \pair{2.09}{0.31} &  \cellhi \pair{2.08}{0.40} &  \cellhi \pair{0.67}{0.05} &  \cellhi \pair{0.59}{0.04} \\
& Regression & L-BFGS            &  \cellhi \pair{0.03}{0.00} &  \cellhi \pair{0.22}{0.01} &  \cellhi \pair{0.61}{0.04} &  \cellhi \pair{0.77}{0.10} &  \cellhi \pair{1.97}{0.38} &  \cellhi \pair{2.08}{0.39} &  \cellhi \pair{0.67}{0.05} &  \pair{0.65}{0.07} \\ \midrule
& Cycle & Adam      &  \pair{0.18}{0.03} &  \pair{0.69}{0.56} &  \pair{1.62}{2.82} &               {\color{gray}>100} &               {\color{gray}>100} &               {\color{gray}>100} &               {\color{gray}>100} &               {\color{gray}>100} \\
& Objective & Adam  &  \pair{0.06}{0.01} &  \pair{0.26}{0.02} &  \cellhi \pair{0.63}{0.07} &  \cellhi \pair{0.81}{0.10} &  \cellhi \pair{1.99}{0.32} &  \cellhi \pair{2.21}{0.32} &  \pair{0.77}{0.05} &  \pair{0.66}{0.07} \\
& Regression & Adam &  \pair{0.22}{0.01} &  \pair{0.28}{0.02} &  \cellhi \pair{0.61}{0.07} &  \cellhi \pair{0.80}{0.10} &  \cellhi \pair{2.07}{0.38} &  \cellhi \pair{2.37}{0.46} &  \pair{0.77}{0.06} &  \pair{0.75}{0.09} \\ \midrule
\multicolumn{3}{r|}{Improvement factor over prior work} & ${\bf 3.3}$ & ${\bf 3.1}$ & ${\bf 3.0}$ & ${\bf 1.8}$ & ${\bf 2.7}$ & ${\bf 1.5}$ & ${\bf 3.0}$ & ${\bf 4.4}$ \\
\bottomrule
\end{tabular}}

\end{table}

\begin{table}[t]
\centering
  \caption{Comparison of $\gL^2$-UVP on the CelebA64 tasks from
    the Wasserstein-2 benchmark by \citet{korotin2021neural}, where
    \abbrev{the gray tags} denote their results.
    I report the mean and standard deviation across 10 trials.
    \textbf{Fine-tuning the amortized prediction with L-BFGS or Adam consistently
    improves the quality of the learned potential.}
    The ConvICNN64 and ResNet potential models are from
    \citet{korotin2021neural}, and \cref{app:conv_potential}
    describes the (non-convex) ConvNet model.
  }
  \label{tab:celeba}
  \vspace{2mm}
\resizebox{0.85\linewidth}{!}{
\begin{tabular}{r@{}ccc|ccc} \toprule
& Amortization loss & Conjugate solver & Potential Model & Early Generator & Mid Generator & Late Generator \\ \midrule
\abbrevtab{W2} & Cycle & {\color{gray}{None}} & ConvICNN64 & 1.7 & 0.5 & 0.25 \\
\abbrevtab{MM} & Objective & {\color{gray}{None}} & ResNet & 2.2 & 0.9 & 0.53 \\
\abbrevtab{MM-R$^\dagger$} & Objective& {\color{gray}{None}} & ResNet & 1.4 & 0.4 & 0.22 \\ \midrule
 & Cycle & None & ConvNet &               {\color{gray}>100} &  \pair{26.50}{60.14} &  \pair{0.29}{0.59} \\
 & Objective & None & ConvNet &               {\color{gray}>100} &    \pair{0.29}{0.15} &  \pair{0.69}{0.90} \\ \midrule
 & Cycle & Adam & ConvNet &  \pair{0.65}{0.02} &    \pair{0.21}{0.00} &  \pair{0.11}{0.04} \\
 & Cycle & L-BFGS & ConvNet &  \cellhi \pair{0.62}{0.01} &    \cellhi \pair{0.20}{0.00} &  \cellhi \pair{0.09}{0.00} \\ \midrule
 & Objective & Adam & ConvNet &  \pair{0.65}{0.02} &    \pair{0.21}{0.00} &  \pair{0.11}{0.05} \\
 & Objective & L-BFGS & ConvNet &  \cellhi \pair{0.61}{0.01} &    \cellhi \pair{0.20}{0.00} &  \cellhi \pair{0.09}{0.00} \\ \midrule
 & Regression & Adam & ConvNet &  \pair{0.66}{0.01} &    \pair{0.21}{0.00} &  \pair{0.12}{0.00} \\
 & Regression & L-BFGS & ConvNet &  \cellhi \pair{0.62}{0.01} &    \cellhi \pair{0.20}{0.00} &  \cellhi \pair{0.09}{0.01} \\ \midrule
\multicolumn{4}{r|}{Improvement factor over prior work} & ${\bf 2.3}$ &  ${\bf 2.0}$ & ${\bf 2.4}$ \\
\bottomrule
\end{tabular}} \\
{\footnotesize
$^\dagger$the \emph{reversed} direction from \citet{korotin2021neural}, \ie
the potential model is associated with the $\beta$ measure}

\end{table}

\section{Demonstrations on synthetic data}
\label{sec:demos}

\begin{figure}[t]
  \centering
  \includegraphics[width=\textwidth]{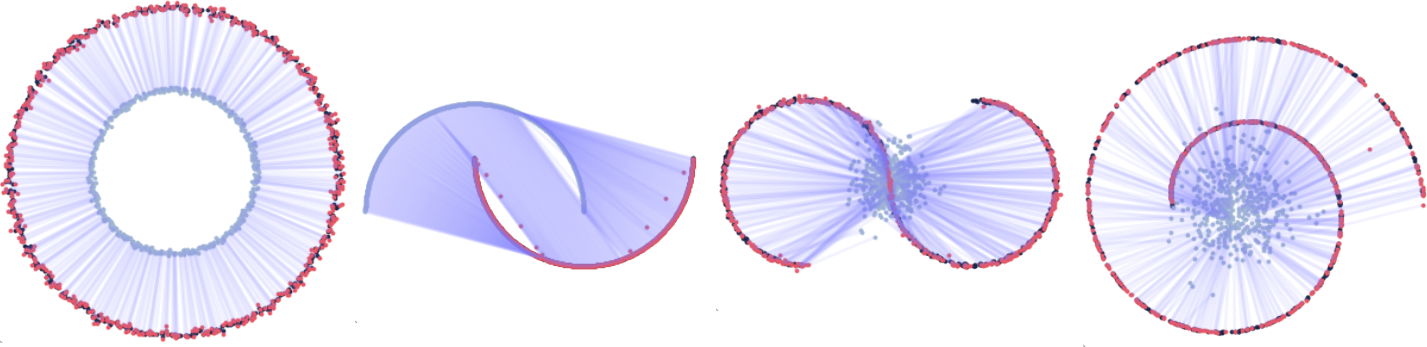} \\
  \cblock{167}{190}{211} Samples from $\beta$ \hspace{2mm}
  \cblock{26}{37}{75} Samples from $\alpha$ \hspace{2mm}
  \cblock{242}{84}{92} Push-forward $(\nabla f^\star)_\# \beta\approx \alpha$ \hspace{2mm}
  \cblock{168}{168}{254} Transport paths
  \caption{Learned transport maps on synthetic settings from \citet{rout2021generative}.}
  \label{fig:rout2021generative}
\end{figure}

\begin{figure}[H]
  \vspace{-2mm}
  \begin{tikzpicture}[every node/.style={align=left,anchor=south west}]
    \node (im) {\includegraphics[width=\textwidth]{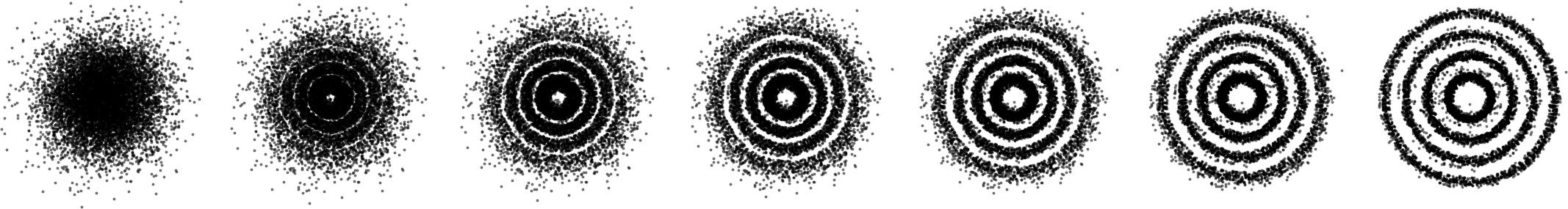} \\
      \includegraphics[width=\textwidth]{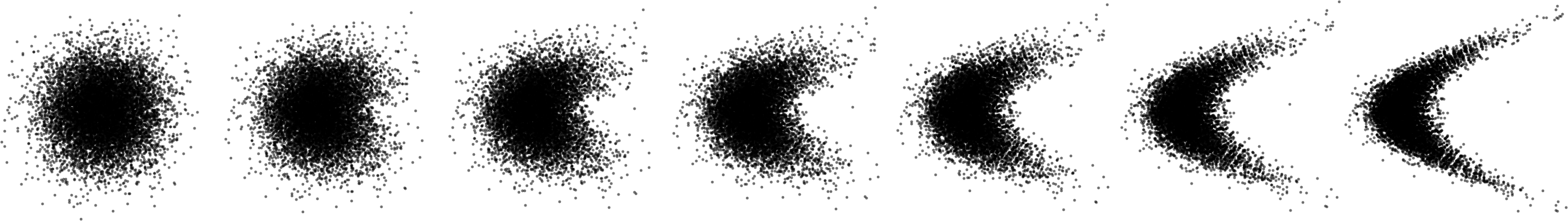} \\
      \includegraphics[width=\textwidth]{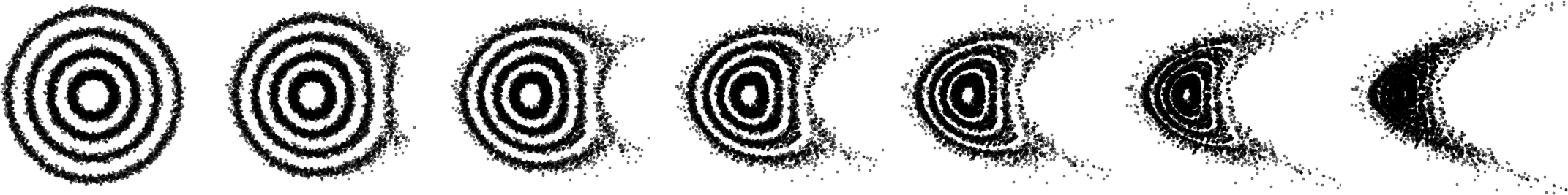}};
    \node[above left=-2mm and -13mm of im] (x) {$\beta$};
    \node[above right=-2mm and -17.5mm of im] (x) {$(\nabla f^\star)_\#\beta$};
    \node[above left=-2mm and -3.70in of im] (x) {\color{gray} $\leftarrow ((1-t)I+t\nabla f^\star)_\#\beta \rightarrow$};
  \end{tikzpicture}
  \caption{Single-block potential flows on synthetic settings
    considered in \citet{huang2020convex}.
  }
  \label{fig:huang2020convex}
  \vspace{-7mm}
\end{figure}

I lastly demonstrate the stability of amortization and fine-tuning
as described in \cref{alg:w2-alg} to learn optimal transport maps between
many 2d synthetic settings considered in the literature.
In all of these settings, I instantiate ICNN and MLP architectures
and use regression-based amortization with L-BFGS fine-tuning.
\Cref{fig:makkuva,fig:grid-wraping,fig:rout2021generative} show the settings considered in
\citet{makkuva2020optimal} and \citet{rout2021generative},
and \cref{fig:makkuva-conj} shows the conjugate objective landscapes.
\Cref{fig:huang2020convex} shows maps learned
on synthetic settings from \citet{huang2020convex}.
\Cref{app:demos} contains more experimental details here.

\begin{remark}
  Optimizing the dual in \cref{eq:kantorovich-dual-parametric}
  is an alternative to the maximum likelihood training
  in \citet{huang2020convex} for potential flows.
  While maximum likelihood training requires the
  density of one of the measures, optimizing the
  dual only requires samples from the measures.
  This makes it easy to compute flows such as the bottom
  one of \cref{fig:huang2020convex}, even though it is
  difficult to specify the density.
  \looseness=-1
\end{remark}

\section{Related work}
\textbf{Numerical conjugation.}
\citet{brenier1989algorithme,lucet1996fast,lucet1997faster,gardiner2013computing,trienis2007computational,jacobs2020fast,vacher2021convex} also use computational methods
for numerical conjugation, which also have applications in solving Hamilton-Jacobi
or Burger's equation via discretizations.
These methods typically consider conjugating univariate and bivariate functions
by discretizing the space, which makes them challenging to apply in
the settings from \citet{korotin2021neural} that we report in
\cref{sec:exp:w2}: we are able to conjugate dual potentials in up to
256-dimensional spaces for the HD tasks and \mbox{12228-dimensional} spaces for the CelebA64 tasks.
Taking a grid-based discretization of the space with 10 locations in each dimension
would result in $(12228)^{10}$ grid points in the CelebA64 task.
\citet{garcia2023fishleg} amortizes the conjugate operation to predict
the natural gradient update, which is related to the amortized
proximal optimization setting in \citet{bae2022amortized}.

\textbf{Learning solutions to OT problems.}
\citet{dinitz2021faster,khodak2022learning,amos2022meta} amortize
and learn the solutions to OT and matching problems by predicting
the optimal duals given the input measures.
These approaches are complementary to this paper as they amortize
the \emph{solution} to the dual in \cref{eq:kantorovich-dual} while this
paper amortizes the conjugate subproblem in \cref{eq:conj} that
is repeatedly computed when solving a single OT problem.

\section{Conclusions, future directions, and limitations}
This paper explores the use of amortization and fine-tuning for
computing convex conjugates.
The methodological insights and amortization perspective may
directly transfer to many other applications and extensions of
Euclidean Wasserstein-2 optimal transport, including for computing
barycenters \citep{korotin2021continuous},
Wasserstein gradient flows \citep{alvarez2021optimizing,mokrov2021large},
or cellular trajectories \citep{bunne2021learning}.
Many of the key amortization and fine-tuning concepts from here will
transfer beyond the Euclidean Wasserstein-2 setting, \eg
the more general $c$-transform arising in
non-Euclidean optimal transport \citep{sei2013jacobian,cohen2021riemannian,rezende2021implicit}
or for the Moreau envelope computation, which can be decomposed
into a term that involves the convex conjugate as
described in
\citet[ex.~11.26]{rockafellar2009variational} and
\citet[sect.~2]{lucet2006fast}.
\looseness=-1

\textbf{Limitations.}
The most significant limitation in the field of estimating
Euclidean Wasserstein-2 optimal transport maps is the lack
of convergence guarantees. The parameter optimization problem
in \cref{eq:kantorovich-dual-parametric} is \emph{always}
non-convex, even when using input-convex neural networks.
I have shown that improved conjugate estimations significantly
improve the stability when the base potential model is
properly set up, but \textbf{all methods are sensitive
to the potential model's hyper-parameters.}
I found that small changes to the activation type or initial learning
rate can cause \emph{no} method to converge.

\begin{figure}[h]
  \vspace{-9mm}
  \centering
  \mbox{\large \hspace{.4in}$(\nabla f)_\#\alpha$ \hspace{.7in} $(\nabla f^\star)_\#\beta$
    \hspace{.6in} Potential $f$ \hspace{.55in} Conjugate $f^*$
    \hspace{.35in}
  } \\
  \includegraphics[width=\textwidth]{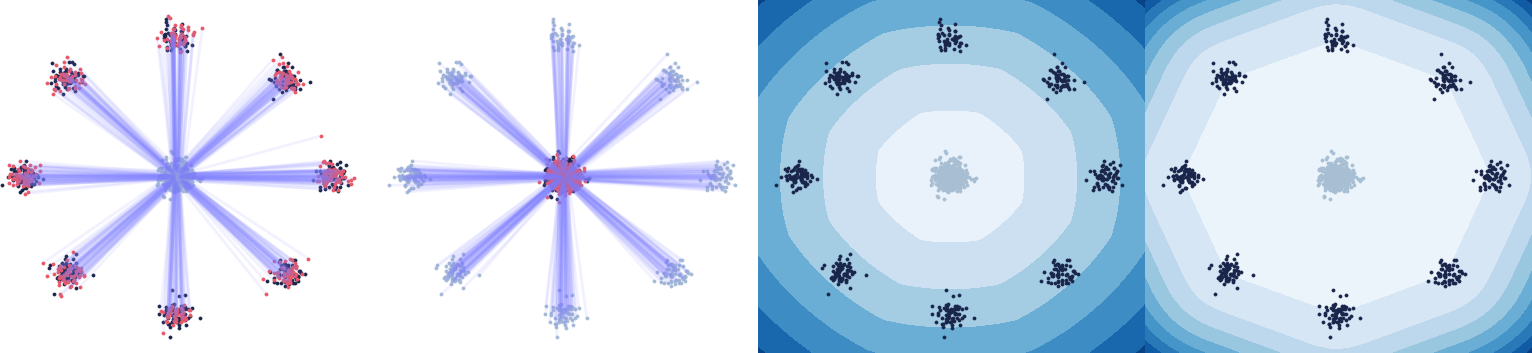} \\[-1mm]
  \includegraphics[width=\textwidth]{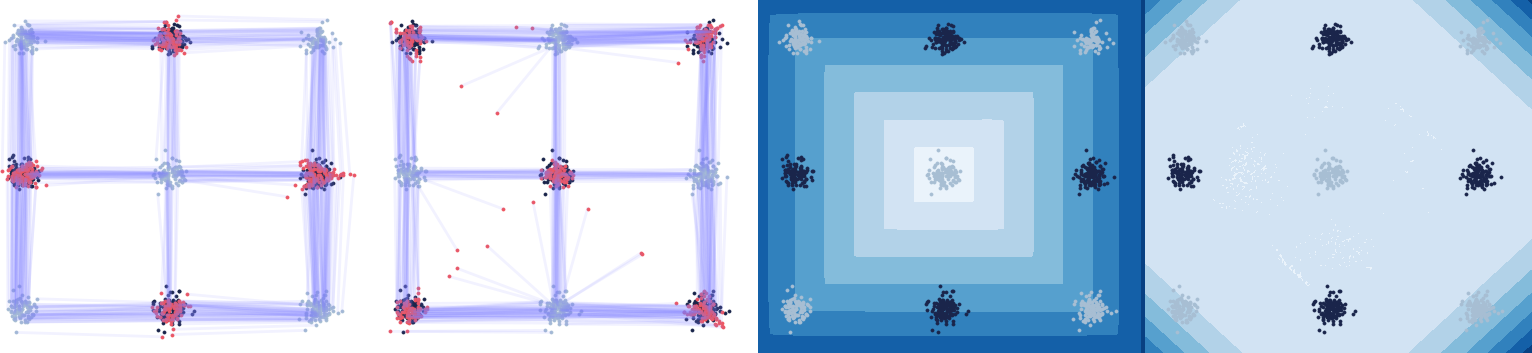} \\
  \cblock{167}{190}{211} Input samples \hspace{2mm}
  \cblock{26}{37}{75} Ground-truth target samples \hspace{2mm}
  \cblock{242}{84}{92} Push-forward samples \hspace{2mm}
  \cblock{168}{168}{254} Transport paths
  \vspace{-2mm}
  \caption{Learned potentials on settings considered in \citet{makkuva2020optimal}.
  }
  \label{fig:makkuva}
\end{figure}

\begin{figure}[H]
  \vspace{-3mm}
  \centering
  \resizebox{\linewidth}{!}{
  \begin{tikzpicture}[every node/.style={align=left,anchor=south west}]
    \node (im) {\includegraphics[width=1.4\textwidth]{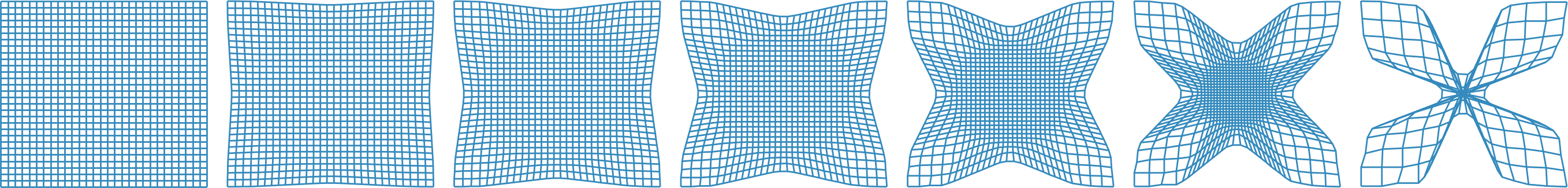}};
    \node[above left=-2mm and -17.5mm of im] (x) {\Large $\gG$};
    \node[above left=-2mm and -5.2in of im] (x) {
      \Large \color{gray} $\leftarrow ((1-t)I+t\nabla f^\star)_\#\gG\rightarrow$};
    \node[above right=-2mm and -25mm of im] (x) {\Large $(\nabla f^\star)_\#\gG$};
  \end{tikzpicture}}
  \vspace{-6mm}
  \caption{Mesh grid $\gG$ warped by the conjugate potential flow $\nabla f^\star$
    from the top setting of \cref{fig:makkuva}.
  }
  \label{fig:grid-wraping}
\end{figure}

\begin{figure}[h]
  \vspace{-2mm}
  \centering
  \begin{tikzpicture}[every node/.style={align=left,anchor=south west}]
    \node (im) {\includegraphics[width=\textwidth]{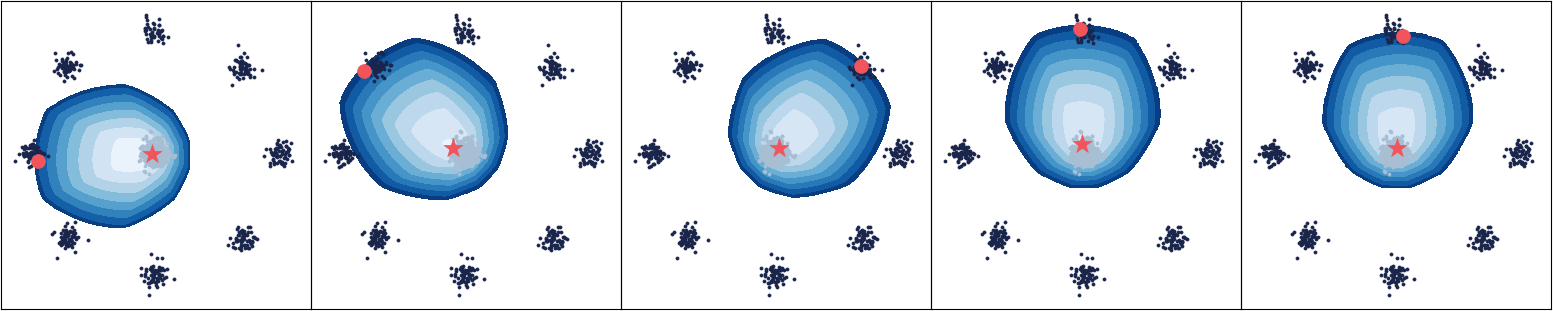}};
    \draw[coral,fill=coral] (2.5mm,31mm) circle[radius=0.6mm,fill=coral] {};
    \node at (3mm,28.3mm) {$y$\hspace{3mm} $\optconj(y)$};
    \draw (8mm,30.5mm) node[star, fill=coral, star point ratio=2.5, inner sep=0.085em] {};
  \end{tikzpicture}
  \vspace{-6mm}
  \caption{Sample conjugation landscapes $J(x; y)$
    of the top setting of \cref{fig:makkuva}.
    The inverse transport map $\nabla_y f^\star(y)=\optconj(y)$
    is obtained by minimizing $J$, which is a convex
    optimization problem.
    The contour shows $J(x; y)$ filtered to not display a
    color for values above $J(y; y)$.
  }
  \label{fig:makkuva-conj}
\end{figure}

\subsection*{Acknowledgments}
I would like to thank Max Balandat,
Ricky Chen,
Samuel Cohen,
Marco Cuturi,
Carles Domingo-Enrich,
Yaron Lipman,
Max Nickel,
Misha Khodak,
Aram-Alexandre Pooladian,
Mike Rabbat,
Adriana Romero Soriano,
Mark Tygert, and
Lin Xiao,
for
insightful comments and discussions.
The core set of tools in Python
\citep{van1995python,oliphant2007python}
enabled this work, including
Hydra \citep{Yadan2019Hydra},
JAX \citep{jax2018github},
Flax \citep{flax2020github},
Matplotlib \citep{hunter2007matplotlib},
numpy \citep{oliphant2006guide,van2011numpy},
and pandas \citep{mckinney2012python}.

{\small
\bibliographystyle{plainnat}
\bibliography{refs}
}

\newpage
\appendix

\section{L-BFGS overview and line search details}
\label{sec:lbfgs}

\begin{algorithm}[t]
  \caption{The Broyden-Fletcher-Goldfarb-Shanno (BFGS) method to solve
    \cref{eq:lbfgs-opt} as presented in \citet[alg.~6.1]{nocedal1999numerical}.}
  \begin{algorithmic}
    \State \textbf{Inputs:} Function $J$ to optimize, initial iterate $x_0$ and
      Hessian approximation $B_0$
    \State $k\leftarrow 0$
    \While {unconverged}
    \State Compute the \emph{search direction} $p_k=-B_k^{-1}\nabla_x J_k(x_k)$
    \State Set $x_{k+1}=x_k+\alpha_k p_k$ where $\alpha_k$ is computed from a \emph{line search}
        from \cref{app:linesearches}
    \State Compute $B_k$ with the update in \cref{eq:bfgs-update}
    \State $k\leftarrow k+1$
    \EndWhile
    \State\Return optimal solution $x_k\approx \breve{x}$
  \end{algorithmic}
  \label{alg:bfgs}
\end{algorithm}

\begin{figure}[t]
  \resizebox{1.\linewidth}{!}{
  \begin{tikzpicture}[every node/.style={align=left,anchor=south west}]
    \node (im) {\includegraphics[height=1.45in]{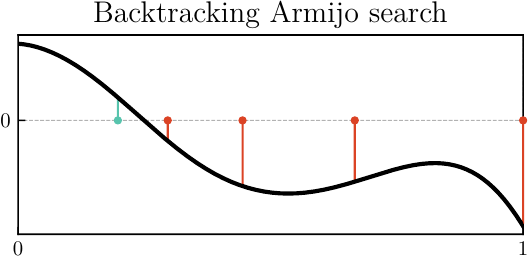}};
    \node at (38mm, 0.5mm) () {$\alpha$};
    \node at (-5mm, 29mm) () {$g(\alpha)$};
    \draw [-{Latex[length=1.5mm,width=1.5mm]}] (75.3mm,21mm) to [out=150,in=30] (52.3mm,21mm);
    \draw [-{Latex[length=1.5mm,width=1.5mm]}] (51.2mm,21mm) to [out=150,in=30] (36.3mm,21mm);
    \draw [-{Latex[length=1.5mm,width=1.5mm]}] (35.3mm,21mm) to [out=150,in=30] (25.7mm,21mm);
    \draw [-{Latex[length=1.5mm,width=1.5mm]}] (24.6mm,21mm) to [out=150,in=30] (18.6mm,21mm);
    \node at (33mm, 24mm) () {\color{gray}Sequentially find $g(\alpha) \geq 0$};
    \node at (15.6mm,15.3mm) () {$\alpha^\star$};
    \draw[thick] (18.05mm,20.85mm) circle (1.5mm);
  \end{tikzpicture}%
  \hspace{-2mm}
  \begin{tikzpicture}[every node/.style={align=left,anchor=south west}]
    \node (im) {\includegraphics[height=1.45in]{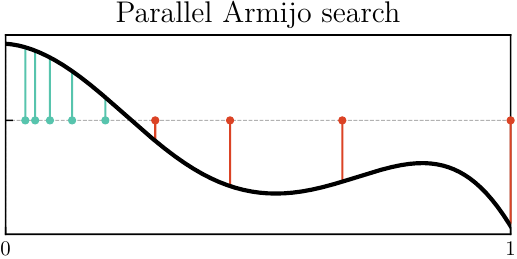}};
    \node at (38mm, 0.5mm) () {$\alpha$};
    \node at (13.5mm, 24mm) () {\color{gray}Evaluate many $g(\alpha)$ in parallel, pick best};
    \node at (13.7mm,15.3mm) () {$\alpha^\star$};
    \draw[thick] (16.3mm,20.85mm) circle (1.5mm);
  \end{tikzpicture}%
  }
  \caption{
    Visualization of backtracking and parallel line searches
    to solve \cref{eq:armijo}.
  }
  \label{fig:linesearch}
\end{figure}

The conjugate optimization problem in \cref{eq:conj} is an unconstrained
convex optimization problem for convex potentials, which is a setting
BFGS \citep{broyden1970convergence,fletcher1970new,goldfarb1970family,shanno1970conditioning}
and L-BFGS \citep{liu1989limited} methods thrive in.
\textbf{The default strong Wolfe line search methods in the
\href{https://github.com/google/jaxopt/blob/418bce35ff7410a86dc5edb64ee5d3716b3bc132/jaxopt/_src/lbfgs.py}{Jax}
and
\href{https://github.com/google/jaxopt/blob/418bce35ff7410a86dc5edb64ee5d3716b3bc132/jaxopt/_src/lbfgs.py}{JaxOpt}
L-BFGS implementations may take a long time to solve a batch of
optimization problems.}
Without efficiently setting the line search method,
some of the Wasserstein-2 benchmark experiments in
\cref{app:w2} that ran in a few hours would have
otherwise taken a \emph{month} to run.
This section provides a brief overview of BFGS methods and
shows that an Armijo line search can be the most efficient at
computing the conjugate.

\subsection{Background on BFGS methods}
\citet[alg.~6.1]{nocedal1999numerical} is a standard reference
for BFGS methods and extensions and the key steps are summarized
in \cref{alg:bfgs} for solving an optimization problem of the form
\begin{equation}
  \breve{x} \in \argmin_x J(x)
  \label{eq:lbfgs-opt}
\end{equation}
where $J: \R^n\rightarrow\R$ is possibly non-convex
and twice continuously differentiable.
The method iteratively finds a solution $\breve{x}$ by
1) maintaining an approximate Hessian around the current iterate,
\ie $B_k\approx \nabla^2 J(x_k)$,
2) computing an approximate \emph{Newton step}
$p_k=-B_k^{-1}\nabla_x J_k(x_k)$ using the approximate Hessian,
3) updating the iterate with $x_{k+1}=x_k+\alpha_k p_k$, where
$\alpha_k$ is found with a \emph{line search}, and
4) updating the Hessian approximation
with the \emph{Sherman–Morrison–Woodbury} formula \citep{woodbury1950inverting,sherman1950adjustment}
\begin{equation}
  B_{k+1}= B_k-\dfrac{B_ks_ss_k^{\top}B_k}{s_k^{\top}B_ks_k}+\dfrac{y_ky_k^\top}{y_k^{\top}s_k},
  \label{eq:bfgs-update}
\end{equation}
where $y_k=\nabla_x J(x_{k+1})-\nabla_x J(x_k)$ and
$s_k=x_{k+1}-x_k$.
Instead of estimating $B_k$ and inverting it in every iteration,
most implementations maintain a direct approximation to the
\emph{inverse} Hessian $H_k\defeq B_k^{-1}$.
The \emph{limited-memory} version of BFGS (L-BFGS) in
\citet{liu1989limited} propose to replace the inverse Hessian
approximation as a \emph{matrix} $H_k$ with the sequence of vectors
$[y_k,s_k]$ defining the updates to $H_k$ and never requires
instantiating the full $n\times n$ approximation.

\newpage

\begin{algorithm}[t]
\caption{Backtracking Armijo line search to solve \cref{eq:armijo}}
\label{alg:backtracking-armijo}
\begin{algorithmic}
\State \textbf{Inputs:} Iterate $x_k$, search direction $p_k$, decay $\tau$, control parameter $c_1$, initial $\alpha_0$
\State $\alpha\leftarrow\alpha_{\rm init}$
\While {$J(x_k+\alpha_jp_k)>J(x_k)+c_1\alpha_jp_k^\top\nabla_xJ(x_k)$}
\State $\alpha\leftarrow \tau\alpha$
\EndWhile
\State\Return $\alpha$ satisfying the Armijo condition in \cref{eq:armijo}.
\end{algorithmic}
\end{algorithm}

\begin{algorithm}[t]
\caption{Parallel Armijo line search to solve \cref{eq:armijo}}
\label{alg:parallel-armijo}
\begin{algorithmic}
  \State \textbf{Inputs:} Iterate $x_k$, search direction $p_k$, decay $\tau$, control parameter $c_1$, initial $\alpha_0$, \#evaluations $M$
\State Compute \emph{candidate step lengths} $\alpha_m=\tau^{-m}$ for $m\in[M]$
\State Evaluate the line search condition $g(\alpha_m)$ from \cref{eq:armijo-condition} in parallel
\If{all $g(\alpha_m) < 0$}
\State \verb!Error: No acceptable step found!
\Else
\State\Return largest $\alpha$ satisfying \cref{eq:armijo},
\ie $\max \alpha_m\ \subjectto\ g(\alpha_m) > 0$
\EndIf
\end{algorithmic}
\end{algorithm}
\subsection{Line searches}
\label{app:linesearches}
The \emph{line search} to find the step size $\alpha_k$ for the
iterate update $x_{k+1}=x_k+\alpha_k p_k$ is often done with:
\begin{enumerate}
\item a \emph{Wolfe} line search \citep{wolfe1969convergence} to satisfy
the conditions:
\begin{equation}
  \begin{aligned}
  J(x_k+\alpha_k p_k) &\leq J(x_k) + c_1 \alpha_k p_k^\top \nabla_x J(x_k) \\
  -p_x^\top\nabla_x J(x_k+\alpha_x p_x) &\leq -c_2 p_x^\top \nabla_x J(x_k) \\
  \label{eq:wolfe}
  \end{aligned}
\end{equation}
where $0<c_1<c_2<1$,
\item a \emph{strong Wolfe} line search to satisfy the conditions:
\begin{equation}
  \begin{aligned}
  J(x_k+\alpha_k p_k)&\leq J(x_k) + c_1 \alpha_k p_k^\top \nabla_x J(x_k) \\
  |p_x^\top\nabla_x J(x_k+\alpha_x p_x)| &\leq c_2 |p_x^\top \nabla_x J(x_k)| \\
  \end{aligned}
  \label{eq:strong-wolfe}
\end{equation}
  This is often found via the \emph{zoom} procedure from
  \citet[algorithm~3.5]{nocedal1999numerical}.
\item an \emph{Armijo} line search \citep{armijo1966minimization}
to satisfy the first condition:
\begin{equation}
  J(x_k+\alpha_k p_k)\leq J(x_k) + c_1 \alpha_k p_k^\top \nabla_x J(x_k).
  \label{eq:armijo}
\end{equation}
For notational simplicity, we can also write the Armijo condition as:
\begin{equation}
  g(\alpha) \defeq f(x_k)+c_1\alpha p_k^\top\nabla f(x_k) - f(x_k+\alpha p_k) \geq 0
  \label{eq:armijo-condition}
\end{equation}
\end{enumerate}

\begin{remark}
The strong Wolfe line search is the most commonly used line search for
L-BFGS as it guarantees that the resulting update to the
Hessian in \cref{eq:bfgs-update} stays positive definite,
but the Armijo line search may be more efficient as it
does not involve re-evaluating the derivative of the objective.
Unfortunately, an iterate obtained by an Armijo line search
may not satisfy the curvature condition $y_k^{\top}s_k>0$
that ensure the Hessian update stays positive definite
while a step satisfying the strong Wolfe
conditions provably does \citep[page~143]{nocedal1999numerical}.
Nonetheless, Armijo searches are still combined with BFGS
and can be \emph{guarded} by only updating the Hessian
approximation if $y_k^{\top}s_k > 0$, or a modification thereof,
as described in \citet{li2001global,wan2012new,fridovich2020approximately} and
\citet[sect.~3.2]{berahas2016multi}.
\label{rmk:armijo-guard}
\end{remark}

\newpage
\subsubsection{The Armijo line search}
The \emph{Armijo line search} can be written as the optimization problem
\begin{equation}
  \alpha_k(x_k, p_k) = \max \alpha\quad\subjectto\quad J(x_k+\alpha p_k) \leq J(x_k)+c_1\alpha p_k^\top\nabla_x J(x_k).
  \label{eq:armijo}
\end{equation}
\Cref{eq:armijo} is typically solved as shown in \cref{alg:backtracking-armijo}
and \cref{fig:linesearch}
by setting a \emph{decay factor} $\tau$
and iteratively decreasing a \emph{candidate step length} $\alpha$
until the condition is satisfied.

When the objective $J$ can be efficiently evaluated in parallel on a GPU,
and when solving many batches of optimization problems concurrently, \eg with \verb!vmap!,
the backtracking Armijo line search described in \cref{alg:backtracking-armijo},
and the Wolfe line search described in \citet[alg.~7.5]{nocedal1999numerical},
are computationally slowed down by serial and conditional operations.
These issues arise from:
1) the sequential nature of the line search, and
2) the fact that the line search may run for a different number of iterations
for every optimization problem in the batch.
Wolfe line searches such as \citet[alg.~7.5]{nocedal1999numerical} have
other conditionals scoping the search interval that cause the line search
to perform potentially different operations for
every optimization problem in the batch.

I propose a \emph{parallel} Armijo line search in
\cref{alg:parallel-armijo}, which is also visualized in \cref{fig:linesearch},
to remove serial and conditional operations to improve the
computation of the line search on the GPU for solving batches of optimization problems.
The key idea is to instantiate many possible step sizes,
evaluate them all at once, and then select the largest $\alpha_m$
satisfying the Armijo condition $g(\alpha_m) \geq 0$.

\begin{remark}
  The parallel line search may unnecessarily evaluate more candidate
  steps sizes than the sequential line search, but on GPU architectures
  this may not be very detrimental to the performance because
  additional parallel function evaluations are computationally cheap.
  Furthermore, when solving a batch of $N$ optimization problems with
  $M$ line search evaluations, \ie when using \verb!vmap! on the line search
  or optimizer, the parallel line search in \cref{alg:parallel-armijo}
  can efficiently evaluate $NM$ candidate step lengths in tandem
  on a GPU and then select the best for each element in the batch.
\end{remark}

\begin{remark}
A potential concern with this parallel line search is that it may not find
a step size satisfying the Armijo condition if $M$ is not set to be low enough.
While this may be a significant issue for when high-precision solves are needed,
I have found in practice for the Euclidean Wasserstein-2 conjugates that
taking $M=10$ line search evaluations frequently finds a solution.
\end{remark}

\subsubsection{Comparing line search methods}

\Cref{tab:lbfgs-line-searches} takes a trained ICNN potential
and isolates the comparison between L-BFGS runtimes to only
changing the linesearch methods.
This is the same optimization procedure and batch size
used for all of the training runs on the Wasserstein-2 benchmark.
Despite the concerns in \cref{rmk:armijo-guard} about the Armijo line
search resulting in slower convergence and an indefinite Hessian
approximations, the Armijo line searches are consistently able to
solve the batch of optimization problems the fastest.

\begin{table}[t]
  \centering
  \caption{
    Runtime and number of L-BFGS iterations for line search methods
    to converge to a solution $\breve{x}$ of \cref{eq:lbfgs-opt}
    for conjugating the trained ICNN potential on the 256-dimensional
    HD benchmark from \cref{sec:exp:w2} with a batch of 1024 samples
    and a tolerance of $\|\nabla J(\breve{x})\|_\infty\leq 0.1$,
    starting from the amortized prediction.
    The Wolfe and Armijo line search
    methods use standard values of $c_1=10^{-4}$ and $c_2=0.9$,
    all backtracking options use a decay factor of $\tau=2/3$
    with $M=15$ evaluations,
    and the runtimes are averaged over 10 trials on an NVIDIA
    Tesla V100 GPU.
    In this setting, Armijo line searches without many conditionals
    or gradient evaluations consistently take the shortest time.
  }
  \label{tab:lbfgs-line-searches}
\begin{tabular}{llcc}
\toprule
Base L-BFGS & Line search & Runtime (ms) &  \# Iterations \\
\midrule
\href{https://github.com/google/jaxopt/blob/418bce35ff7410a86dc5edb64ee5d3716b3bc132/jaxopt/_src/lbfgs.py}{Jax} & Strong Wolfe Zoom (default) &    $4803.40$ &  $6.21$ \\
& Backtracking Armijo          &     $156.69$ &  $8.56$ \\
& Parallel Armijo              &  \cellhi $119.41$ &  $8.56$ \\ \midrule
\href{https://github.com/google/jaxopt/blob/418bce35ff7410a86dc5edb64ee5d3716b3bc132/jaxopt/_src/lbfgs.py}{JaxOpt} & Strong Wolfe Zoom (default) &     $776.52$ &  $7.78$ \\
& Backtracking Strong Wolfe &     $233.39$ &  $7.97$ \\
& Backtracking Wolfe        &     $225.30$ &  $8.65$ \\
& Backtracking Armijo       &     $154.90$ &  $8.58$ \\
\bottomrule
\end{tabular}
\vspace*{-3mm}
\end{table}

\newpage
\section{Model definitions and pretraining}
\label{app:models}
All of the potential and conjugate amortization models in this paper can be implemented
in $\approx\!30$--$50$ lines of readable Jax code with Flax \citep{flax2020github}.
They are included in the \verb!w2ot/models! directory of the code
\href{http://github.com/facebookresearch/w2ot}{here}, and
reproduced here to precisely define them.

\subsection{Pretraining and initialization}
Following \citet{korotin2021neural}, every experimental setting has a pre-training
phase so that the potentials and amortization maps approximate the identity mapping,
\ie $\nabla_x f_\theta(x)\approx x$ and $\tilde x_\varphi(y)\approx y$.
\looseness=-1

\subsection{{\tt InitNN}: Non-convex neural network amortization model $\tilde x_\varphi$}
\label{app:init_nn}

\begin{remark}
  The passthrough on line 18 is helpful for
  learning an identity initialization.
\end{remark}

\begin{lstlisting}[language=Python]
class InitNN(nn.Module):
    dim_hidden: Sequence[int]
    act: str = 'elu'

    @nn.compact
    def __call__(self, x):
        assert x.ndim == 2
        n_input = x.shape[-1]

        act_fn = layers.get_act(self.act)

        z = x
        for n_hidden in self.dim_hidden:
            Wx = nn.Dense(n_hidden, use_bias=True)
            z = act_fn(Wx(z))

        Wx = nn.Dense(n_input, use_bias=True)
        z = x + Wx(z) # Encourage identity initialization.

        return z
\end{lstlisting}

\newpage
\subsection{{\tt ICNN}: Input-convex neural network potential $f_\theta$}
\label{app:icnn}
\verb!actnorm! is the \emph{activation normalization} layer from \citet{kingma2018glow},
which was also used in the ICNN potentials in \citet{huang2020convex} and
normalizes the activations at initialization to follow a normal distribution.

\begin{remark}
  \label{icnn:remark1}
  Applying an activation to the output on line 41 is helpful to
  lower-bound the otherwise unconstrained potential and
  adds stability to the training.
\end{remark}

\begin{remark}
  \label{icnn:remark2}
  The final quadratic on line 46 makes it easy to initialize the
  potential to the identity.
\end{remark}

\begin{remark}
  This ICNN does \textbf{not} use the quadratic activations proposed in
  \citet[Appendix~B.1]{korotin2019wasserstein}. While I did not heavily
  experiment with them, \cref{tab:hd} shows that this
  ICNN architecture without the quadratic activations performs better
  than the results originally reported in \citet{korotin2021neural}
  which use an ICNN architecture with the quadratic activations.
\end{remark}

\begin{lstlisting}[language=Python]
class ICNN(nn.Module):
    dim_hidden: Sequence[int]
    act: str = 'elu'
    actnorm: bool = True

    def setup(self):
        kernel_init = nn.initializers.variance_scaling(
            1., "fan_in", "truncated_normal")
        num_hidden = len(self.dim_hidden)

        w_zs = list()
        for i in range(1, num_hidden):
            w_zs.append(layers.PositiveDense(
                self.dim_hidden[i], kernel_init=kernel_init))
        w_zs.append(layers.PositiveDense(1, kernel_init=kernel_init))
        self.w_zs = w_zs

        w_xs = list()
        for i in range(num_hidden):
            w_xs.append(nn.Dense(
                self.dim_hidden[i], use_bias=True,
                kernel_init=kernel_init))

        w_xs.append(nn.Dense(1, use_bias=True, kernel_init=kernel_init))
        self.w_xs = w_xs


    @nn.compact
    def __call__(self, x):
        assert x.ndim == 2
        n_input = x.shape[-1]
        act_fn = layers.get_act(self.act)

        z = act_fn(self.w_xs[0](x))
        for Wz, Wx in zip(self.w_zs[:-1], self.w_xs[1:-1]):
            z = Wz(z) + Wx(x)
            if self.actnorm:
                z = layers.ActNorm()(z)
            z = act_fn(z)

        y = act_fn(self.w_zs[-1](z) + self.w_xs[-1](x))
        y = jnp.squeeze(y, -1)

        log_alpha = self.param(
            'log_alpha', nn.initializers.constant(0), [])
        y += jnp.exp(log_alpha)*0.5*utils.batch_dot(x, x)

        return y
\end{lstlisting}

\newpage
\subsection{{\tt PotentialNN}: Non-convex neural network (MLP) potential $f_\theta$}
\label{app:potential_nn}
\begin{remark}
  Consistent with \cref{icnn:remark1,icnn:remark2}, applying an activation to the
  last layer (line 18) and adding a final quadratic term (line 24)
  helps this non-convex potential model too.
\end{remark}

\begin{lstlisting}[language=Python]
class PotentialNN(nn.Module):
    dim_hidden: Sequence[int]
    act: str = 'elu'

    @nn.compact
    def __call__(self, x):
        assert x.ndim == 2
        n_input = x.shape[-1]

        act_fn = layers.get_act(self.act)

        z = x
        for n_hidden in self.dim_hidden:
            Wx = nn.Dense(n_hidden, use_bias=True)
            z = act_fn(Wx(z))

        Wx = nn.Dense(1, use_bias=True)
        z = act_fn(Wx(z))

        z = jnp.squeeze(z, -1)

        log_alpha = self.param(
            'log_alpha', nn.initializers.constant(0), [])
        z += 0.5*jnp.exp(log_alpha)*utils.batch_dot(x, x)

        return z
\end{lstlisting}

\newpage
\subsection{{\tt ConvPotential}: Non-convex convolutional potential $f_\theta$}
\label{app:conv_potential}

\begin{remark}
  I was not able to easily add batch normalization \citep{ioffe2015batch}
  to this potential.
  In contrast to standard use cases of batch normalization
  that only call into a batch-normalized model once over samples from
  a single distribution,
  the dual objective in \cref{eq:kantorovich-dual-parametric} calls
  into the potential multiple times to estimate
  $\E_{x\sim\alpha} f_\theta(x)$ and $\E_{y\sim\beta} f_\theta(\optconj(y))$,
  which also involve internally solving the conjugate optimization
  problem in \cref{eq:conj} to obtain $\optconj$.
  This makes it not clear what training and evaluation statistics
  batch normalization should use when computing the dual objective.
  One choice could be to only use the statistics
  induced from the samples $x\sim\alpha$.
\end{remark}

\begin{lstlisting}[language=Python]
class ConvPotential(nn.Module):
    act: str = 'elu'

    mean = jnp.array([0.485, 0.456, 0.406])
    std = jnp.array([0.229, 0.224, 0.225])

    @nn.compact
    def __call__(self, x):
        assert x.ndim == 2 # Images should be flattened
        num_batch = x.shape[0]

        x_flat = x # Save for taking the quadratic at the end.

        # Reshape and renormalize
        x = x.reshape(-1, 3, 64, 64).transpose(0, 2, 3, 1)
        x = (x + 1.)/2.
        x = (x-self.mean) / self.std
        y = x

        act_fn = layers.get_act(self.act)

        conv = nn.Conv(128, kernel_size=[4,4], strides=2)
        y = act_fn(conv(y))

        conv = nn.Conv(128, kernel_size=[4,4], strides=2)
        y = act_fn(conv(y))

        conv = nn.Conv(256, kernel_size=[4,4], strides=2)
        y = act_fn(conv(y))

        conv = nn.Conv(512, kernel_size=[4,4], strides=2)
        y = act_fn(conv(y))

        conv = nn.Conv(1024, kernel_size=[4,4], strides=2)
        y = act_fn(conv(y))

        conv = nn.Conv(
            1, kernel_size=[2,2], padding='VALID', strides=1)
        y = act_fn(conv(y))
        y = y.squeeze([1,2,3])

        assert y.shape == (num_batch,)

        log_alpha = self.param(
            'log_alpha', nn.initializers.constant(0), [])
        y += 0.5*jnp.exp(log_alpha)*utils.batch_dot(x_flat, x_flat)

        return y
\end{lstlisting}
\vfill

\section{Additional Wasserstein-2 benchmark experiment details}
\label{app:w2}

\subsection{Hyper-parameters}
\label{app:w2hypers}

\Cref{tab:w2hd-hypers,tab:w2celeba-hypers} detail the main hyper-parameters
for the Wasserstein-2 benchmark experiments. I tried to keep these
consistent with the choices from \citet{korotin2021neural}, \eg using the
same batch sizes, number of training iterations, and hidden layer sizes for
the potential.

All experiments use the same settings for the conjugate solvers:
The conjugate solvers stop early if all dimensions of the
iterates change by less than 0.1, and otherwise run for a maximum of
100 iterations.
The line search parameters for the parallel Armijo search in \cref{alg:parallel-armijo}
for L-BFGS are to decay the steps with a base of $\tau=1.5$ and
to search $M=10$ step sizes.
With the Adam conjugate solver, I use the default $\beta=[0.9, 0.999]$
with an initial learning rate of 0.1 with a cosine annealing schedule
to decrease it to $10^{-5}$.

\begin{table}[H]
  \centering
\caption{Hyper-parameters for the $D$-dimensional Wasserstein-2 benchmark experiments}
\label{tab:w2hd-hypers}
\begin{tabular}{rl}\toprule
Name & Value \\\midrule
potential model $f_\theta$ & {\tt ICNN} or {\tt PotentialNN} \\
$f_\theta$ hidden layer sizes & [$\max(2D,64)$, $\max(2D,64)$, $\max(D,32)$] \\
conjugate amortization model $\tilde x_\varphi$ & {\tt InitNN(dim\_hidden=[512, 512])} \\
activation functions & ELU \citep{clevert2015fast} \\
\# training iterations & 250000 \\
optimizer & Adam with \href{https://optax.readthedocs.io/en/latest/api.html#optax.cosine_decay_schedule}{cosine annealing} ($\alpha$=1e-4) \\
initial learning rate & 5e-4 \\
Adam $\beta$ & [0.5, 0.5] \\
batch size & 1024 \\
\bottomrule
\end{tabular}
\end{table}

\begin{table}[H]
  \centering
\caption{Hyper-parameters for the CelebA64 Wasserstein-2 benchmark experiments}
\label{tab:w2celeba-hypers}
\begin{tabular}{rl}\toprule
Name & Value \\\midrule
potential model $f_\theta$ & {\tt ConvPotential} \\
conjugate amortization model $\tilde x_\varphi=\nabla g_\varphi$ & Gradient of {\tt ConvPotential} \\
activation functions & ELU \citep{clevert2015fast} \\
number of training iterations & 50000 \\
optimizer & Adam with \href{https://optax.readthedocs.io/en/latest/api.html#optax.cosine_decay_schedule}{cosine annealing} ($\alpha$=1e-4) \\
initial learning rate & 1e-3 \\
Adam $\beta$ & [0.5, 0.5] \\
batch size & 64 \\
\bottomrule
\end{tabular}
\end{table}

\newpage
\subsection{Convergence of L-BFGS and Adam for
  solving the conjugate}
\label{app:convergence}

\begin{figure}[H]
  \centering
  \includegraphics[width=\textwidth]{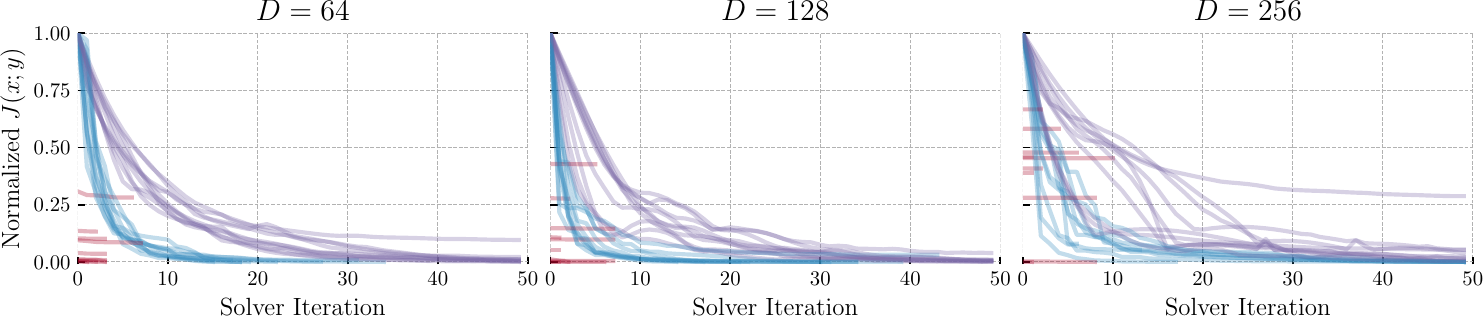}
  \cblock{52}{138}{189} L-BFGS \hspace{2mm}
  \cblock{166}{6}{40} Amortization + L-BFGS \hspace{2mm}
  \cblock{128}{114}{179} Adam
  \caption{Conjugate solver convergence on the HD benchmarks with a NN potential.}
  \label{fig:conj-convergence-nn-hd}
\end{figure}

\begin{figure}[H]
  \centering
  \includegraphics[width=\textwidth]{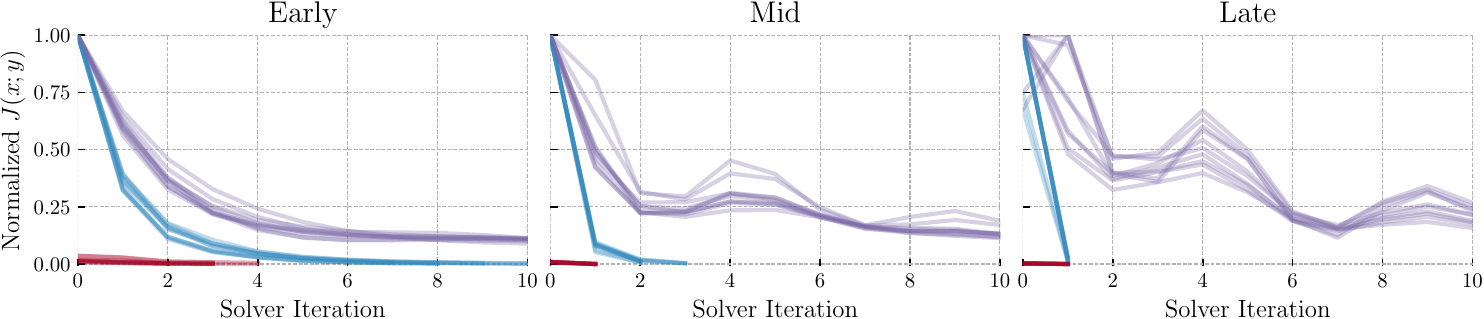}
  \cblock{52}{138}{189} L-BFGS \hspace{2mm}
  \cblock{166}{6}{40} Amortization + L-BFGS \hspace{2mm}
  \cblock{128}{114}{179} Adam
  \caption{Conjugate solver convergence on the CelebA64 benchmark}
  \label{fig:conj-convergence-nn-celeba}
\end{figure}

\Cref{fig:conj-convergence-nn-hd}
shows that with a non-convex potential, many
of the initial amortized predictions are suboptimal and difficult
for the L-BFGS to improve upon.
This indicates that the amortized predictions may
be in parts of the space that are difficult to recover from
and suggests a future avenue of work better characterizing
and recovering from this behavior.
L-BFGS converges fast to an optimal solution in \cref{fig:conj-convergence-nn-celeba}
while Adam often gets stuck at suboptimal solutions.

\subsection{Additional runtimes and conjugation information}
\label{app:w2:runtimes}

\Cref{tab:hd-info,tab:celeba-info} contain additional experimental
information with the:
\begin{enumerate}
\item \textbf{wall-clock time} for the entire training run
  measured on an NVIDIA Tesla V100 GPU,
\item \textbf{number of conjugation iterations}
  from \cref{alg:conjugate} for the conjugate solver to converge
  at the end of training after warm-starting it from the
  conjugate amortization model's prediction, and
\item \textbf{runtime for the conjugate solver} to converge on a batch
  of instances, set to the batch size used during training,
  \ie 1024 for the HD benchmark and 64 for the CelebA64 benchmark.
\end{enumerate}

These results give an idea of how much additional time is spent fine-tuning.
On the HD benchmark, fine-tuning takes between
$\approx\!10$--$50$ms per batch.
The overall wall clock time may take $\approx\!2$--$3$ times
longer than the training runs without fine-tuning,
but are able to find significantly better solutions.
On the CelebA64 benchmarks, the conjugation time impacts the
overall runtime even less because, especially in the
``Mid'' and ``Late'' settings as the transport maps here are
close to being the identity mapping and are easy to conjugate.

\begin{remark}
  Some settings immediately diverged to an irrecoverable state
  providing a $\gL^2$-UVP of $10^9$, including runs using the
  objective-based and cycle amortization losses without fine-tuning.
  I early-stopped those experiments and do not report the runtimes
  or conjugation times here, as the few minutes that the objective-based
  amortization experiments took to diverge is not very interesting
  or comparable to the times of the experiments that converged.
\end{remark}

\begin{table}[t]
  \caption{Additional runtime and conjugation information for the HD benchmark.
    These report the median time from the converged runs.}
  \label{tab:hd-info}
  \centering
\resizebox{1.\linewidth}{!}{
\begin{tabular}{lll|lll|lll|lll}
\toprule
& & & \multicolumn{3}{c|}{Runtime (hours)} & \multicolumn{3}{c|}{Final conjugation iter} & \multicolumn{3}{c}{Conj runtime (seconds)} \\
& Amortization loss & Conjugate solver & $n=64$  &   $n=128$ &    $n=256$ & $n=64$  &   $n=128$ &    $n=256$ & $n=64$  &   $n=128$ &    $n=256$ \\
\midrule
ICNN & Cycle & {\color{gray}None}     &  1.10 &  1.24 &   1.58 & & & & \\ \midrule
ICNN & Cycle & L-BFGS                 &  4.39 &  7.76 &  21.14 & 14.32 &  19.38 &  70.23 &  0.05 &  0.10 &  0.29 \\
ICNN & Objective & L-BFGS             &  2.69 &  4.70 &  13.86 & 5.43 &   6.96 &   8.13 &  0.02 &  0.05 &  0.13 \\
ICNN & Regression & L-BFGS            &  2.74 &  4.30 &  12.53 & 5.64 &   6.34 &   8.73 &  0.02 &  0.04 &  0.11 \\ \midrule
ICNN & Cycle & Adam      &  2.09 &  2.80 &  0.86 & 37.23 &  51.34 &  97.21 &  0.02 &  0.03 &  0.06 \\
ICNN & Objective & Adam  &  2.23 &  2.99 &  4.94 & 29.44 &  38.81 &  55.05 &  0.02 &  0.03 &  0.05 \\
ICNN & Regression & Adam &  2.09 &  2.86 &  4.81 & 29.51 &  38.20 &  54.21 &  0.02 &  0.03 &  0.06 \\ \midrule
NN & Objective & L-BFGS             &  2.31 &  3.34 &  6.05 & 5.05 &   6.50 &   5.77 &  0.02 &  0.03 &  0.05 \\
NN & Regression & L-BFGS            &  2.28 &  3.21 &  5.76 & 5.21 &   4.72 &   5.01 &  0.02 &  0.02 &  0.05 \\ \midrule
NN & Objective & Adam  &  1.61 &  2.18 &  3.70 & 27.84 &   36.81 &  45.82 &  0.01 &  0.02 &  0.04 \\
NN & Regression & Adam &  1.77 &  2.51 &  3.79 & 28.28 &   31.46 &  40.31 &  0.01 &  0.02 &  0.04 \\
\bottomrule
\end{tabular}}
\end{table}

\begin{table}[t]
  \caption{Additional runtime and conjugation information for the CelebA64 benchmark}
  \label{tab:celeba-info}
  \centering
\resizebox{1.\linewidth}{!}{
\begin{tabular}{llll|lll|lll|lll}
\toprule
  & & & & \multicolumn{3}{c|}{Runtime (hours)} &
    \multicolumn{3}{c|}{Final conj iter} & \multicolumn{3}{c}{Conj runtime (seconds)} \\
Amortization loss & Conjugate solver & Model & Direction & Early & Mid & Late &
    Early & Mid & Late & Early & Mid & Late \\ \midrule
Objective & {\color{gray}None} & Conv & Forward     &  3.28 &  3.41 &  3.28 &   \\
Cycle & {\color{gray}None} & Conv & Forward         &  0.94 &  4.17 &  4.17 &   \\ \midrule
Cycle & Adam & Conv & Forward         &  5.44 &  4.78 &  3.57 &  18.23 &  2.02 &  2.80 &  0.15 &  0.11 &  0.04 \\
Cycle & L-BFGS & Conv & Forward       &  6.38 &  3.79 &  3.74 &   5.18 &  2.00 &  2.00 &  0.22 &  0.04 &  0.04 \\ \midrule
Objective & Adam & Conv & Forward     &  5.40 &  4.85 &  3.47 &  19.87 &  1.83 &  1.79 &  0.17 &  0.13 &  0.03 \\
Objective & L-BFGS & Conv & Forward   &  6.09 &  3.86 &  3.70 &   4.48 &  2.01 &  2.00 &  0.21 &  0.06 &  0.04 \\ \midrule
Regression & Adam & Conv & Forward    &  5.44 &  4.86 &  3.51 &  22.12 &  2.84 &  1.01 &  0.19 &  0.14 &  0.02 \\
Regression & L-BFGS & Conv & Forward  &  5.96 &  3.77 &  3.66 &   4.55 &  2.01 &  2.02 &  0.22 &  0.04 &  0.04 \\
\bottomrule
\end{tabular}}
\end{table}

\section{Additional 2d demonstration experiment details}
\label{app:demos}

\begin{figure}[t]
  \centering
  \begin{tikzpicture}[every node/.style={align=left,anchor=south west}]
    \node (im) {\includegraphics[width=\textwidth]{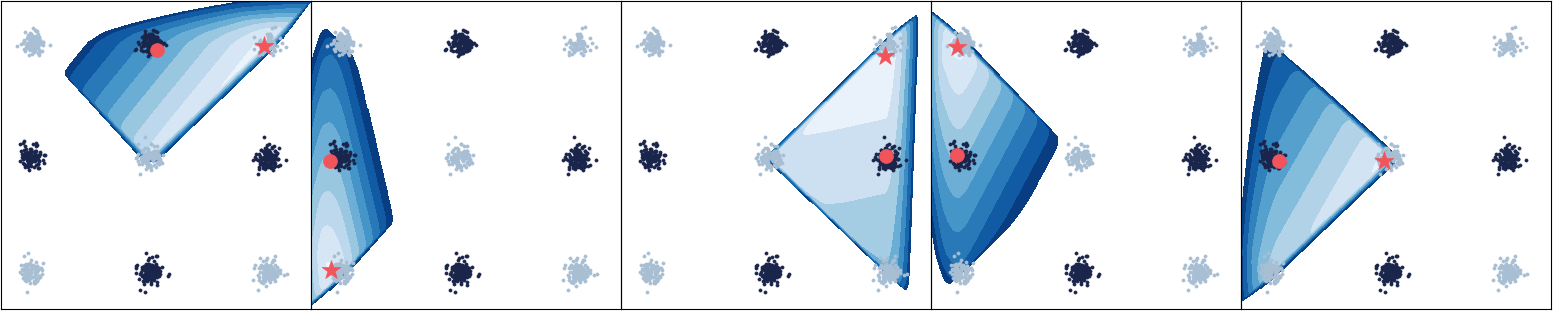}};
    \draw[coral,fill=coral] (2.5mm,31mm) circle[radius=0.6mm,fill=coral] {};
    \node at (3mm,28.3mm) {$y$\hspace{3mm} $\optconj(y)$};
    \draw (8mm,30.5mm) node[star, fill=coral, star point ratio=2.5, inner sep=0.085em] {};
  \end{tikzpicture}
  \caption{Sample conjugation landscapes $J(x; y)$
    of the bottom setting of \cref{fig:makkuva}.
    The inverse transport map $\nabla_y f^\star(y)=\optconj(y)$
    is obtained by minimizing $J$, which is a convex
    optimization problem.
    The contour shows $J(x; y)$ filtered to not display a
    color for values above $J(y; y)$.
  }
  \label{fig:makkuva-conj-2}
\end{figure}

\Cref{tab:synthetic-hypers} details the main hyper-parameters for
the synthetic benchmark experiments, and
\cref{fig:makkuva-conj-2} shows additional conjugation landscapes.

\begin{remark}
  I found leaky ReLU activations on the potential model to work better
  in these low-dimensional settings than ELU activations,
  which work better in the HD benchmark settings.
  I do not have a strong explanation for this but found the LReLU
  capable of performing sharper transitions in the space,
  \eg the sharp boundaries shown in \cref{fig:makkuva}.
  One reason that the ELU potentials could perform better
  on the benchmark settings is that the ground-truth
  transport maps in the benchmark, described in
  \citet[Appendix~B.1]{korotin2021neural},
  use an ICNN with CELU activations \citep{barron2017continuously}
  which may be easier to recover with potential models that
  use ELU activations.
\end{remark}

I trained convex and non-convex potentials on every synthetic setting
and show the results from the best-performing potential model, which are:
\begin{itemize}
\item \citet{makkuva2020optimal}: an ICNN.
  This setting originally considered convex potentials, and the non-convex
  potentials I tried training on these settings diverged,
\item \citet{rout2021generative}: a non-convex potential (an MLP).
  This setting also originally considered an MLP and I couldn't
  find an ICNN that accurately transports between the highly curved
  and concentrated parts of the measures.
\item \citet{huang2020convex}: a non-convex potential (an MLP).
  \textbf{In contrast to the ICNNs originally used,
  I found that an MLP works better when learned with the OT dual.}
  Almost every setting in \citet{huang2020convex} requires composing
  multiple blocks of ICNNs, which means the flow will not necessarily
  be the optimal transport flow, while the non-convex MLP potential
  I am using here estimates the \emph{optimal} transport map between the measures.
\end{itemize}

All of the synthetic settings use the L-BFGS conjugate solver set to
obtain slightly higher precision solves than in the Wasserstein-2
benchmark.
The conjugate solver stops early if all dimensions of the
iterates change by less than 0.001, and otherwise run for a maximum of
100 iterations.
The line search parameters for the parallel Armijo search in \cref{alg:parallel-armijo}
for L-BFGS are to decay the steps with a base of $\tau=1.5$ and
to search $M=30$ step sizes.

\begin{table}[t]
  \centering
\caption{Hyper-parameters for the synthetic experiments}
\label{tab:synthetic-hypers}
\begin{tabular}{rl}\toprule
Name & Value \\\midrule
potential model $f_\theta$ & {\tt ICNN} or {\tt PotentialNN} \\
$f_\theta$ hidden layer sizes & [128, 128] \\
conjugate amortization model $\tilde x_\varphi$ & {\tt InitNN(dim\_hidden=[512, 512])} \\
activation functions & Leaky ReLU with slope 0.2 \\
\# training iterations & 50000 \\
optimizer & Adam with \href{https://optax.readthedocs.io/en/latest/api.html#optax.cosine_decay_schedule}{cosine annealing} ($\alpha$=1e-4) \\
initial learning rate & 5e-4 \\
Adam $\beta$ & [0.5, 0.5] \\
batch size & 10000 \\
\bottomrule
\end{tabular}
\end{table}

\subsection{Non-convex regions in the learned potentials}
\emph{Brenier's theorem} \citep{brenier1991polar} shows
that the known Wasserstein-2 optimal transport map associated with the
negative inner product cost is the gradient of a convex function, \ie
$\optprimal(x) = \nabla_x \optdual(x)$.
Because of this, optimizing over convex potentials is theoretically
nice and also results in a convex and easy conjugate optimization
problem in \cref{eq:conj} to compute $f^\star$.
The input-convex property is usually enforced by constraining all of
the weights of the network to be positive in every layer except the first.
Unfortunately, in practice, the positivity constraints of
a convex potential may be prohibitive and not easy to optimize over
and result in sub-optimal transport maps.
In other words, the parameter optimization problem over the input-convex
model is still non-convex and may be exacerbated by the input-convex constraints.
Due to these limitations, non-convex potentials are appealing as their
parameter space is less constrained and may therefore be easier to
search over. And in practice, this has been shown to be true, \eg
the main results in \cref{tab:hd} show that a non-convex potential
significantly outperforms the convex potential.
However, non-convex potentials can result in non-convex conjugate
optimization problems in \cref{eq:conj} that can cause
significant numerical instabilities and an exploding upper-bound on
the dual objective.

\begin{figure}[H]
  \centering
  Interpolation from a non-convex potential {\color{gray}(an MLP)} \\
  \resizebox{1.\linewidth}{!}{
  \fbox{\begin{tikzpicture}[every node/.style={align=left,anchor=south west}]
    \node (im) {\includegraphics[width=\textwidth]{fig/maf-moons-interp.png}\vspace{-2mm}};
    \node[above left=-3mm and -13mm of im] (x) {$\beta$};
    \node[above right=-3mm and -17.5mm of im] (x) {$(\nabla f^\star)_\#\beta$};
    \node[above left=-3mm and -3.70in of im] (x) {\color{gray} $\leftarrow ((1-t)I+t\nabla f^\star)_\#\beta \rightarrow$};
  \end{tikzpicture}}} \\[3mm]
  Interpolation from a convex potential {\color{gray}(an ICNN)} \\
  \resizebox{1.\linewidth}{!}{
  \fbox{\begin{tikzpicture}[every node/.style={align=left,anchor=south west}]
    \node (im) {\includegraphics[width=\linewidth]{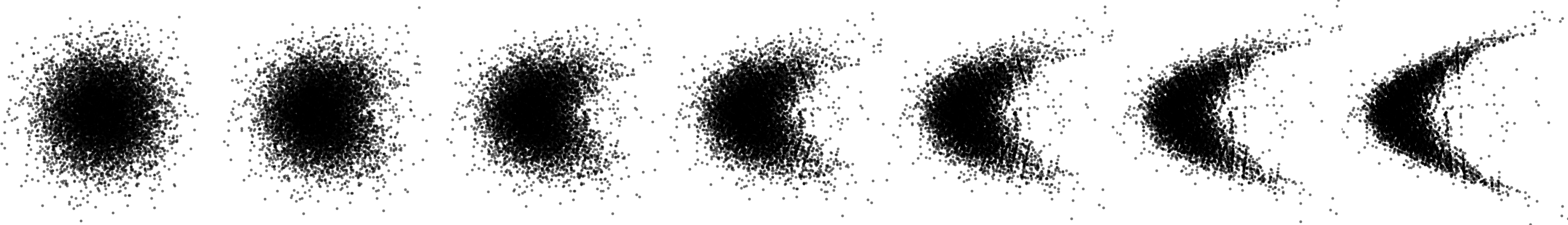}\vspace{-2mm}};
    \node[above left=-3mm and -13mm of im] (x) {$\beta$};
    \node[above right=-3mm and -17.5mm of im] (x) {$(\nabla f^\star)_\#\beta$};
    \node[above left=-3mm and -3.70in of im] (x) {\color{gray} $\leftarrow ((1-t)I+t\nabla f^\star)_\#\beta \rightarrow$};
  \end{tikzpicture}}} \\[3mm]
  \begin{minipage}{0.35\linewidth}
    \centering
    Non-convex potential contours \\
    \includegraphics[width=\linewidth]{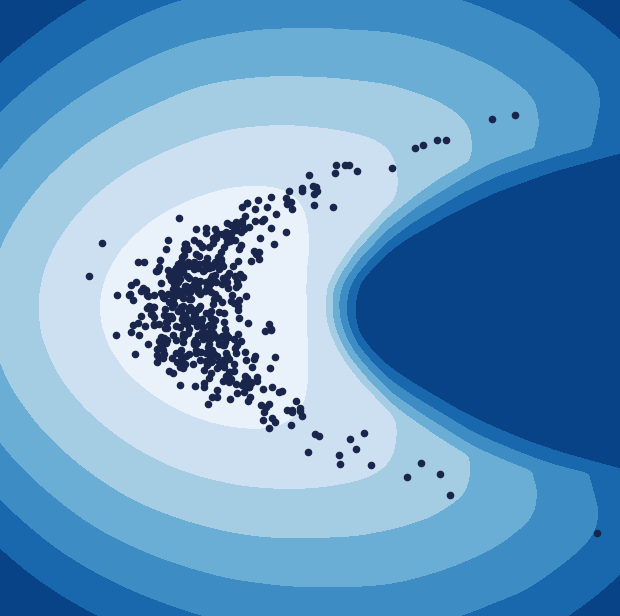}
  \end{minipage} \hspace{5mm}
  \begin{minipage}{0.35\linewidth}
    \centering
    Convex potential contours \\
    \includegraphics[width=\linewidth]{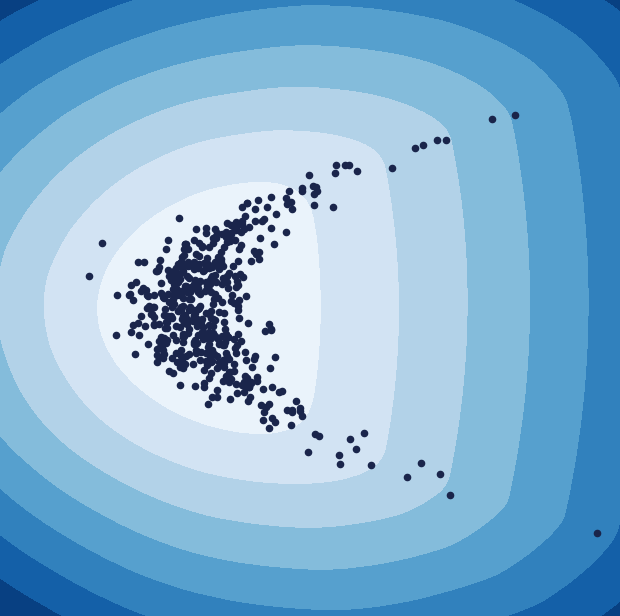}
  \end{minipage}
  \caption{Convex and non-convex potentials trained on the same transport task.}
  \label{fig:mlp-vs-icnn}
\end{figure}

\Cref{fig:mlp-vs-icnn} illustrates a small non-convex region arising in
a learned non-convex potential. While the non-convex region
mostly does not impact the transport map in this case,
they can easily blow up and make the dual
optimization problem challenging.
In contrast, the ICNN-based convex potential provably retains
convexity and keeps this region nicely flat, but the constraints
on the parameter space may hinder the performance.
\newpage

\subsection{Interpolations on synthetic settings from \citet{rout2021generative}}
\fbox{\includegraphics[width=0.9\textwidth]{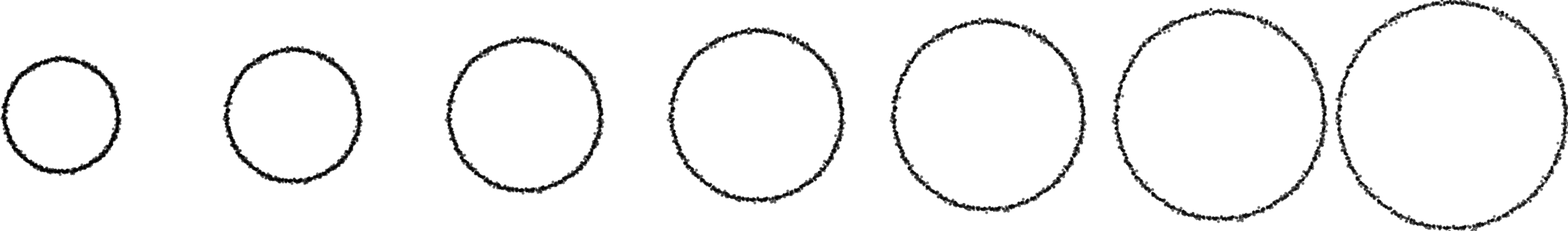}} \\[2mm]
\fbox{\includegraphics[width=0.9\textwidth]{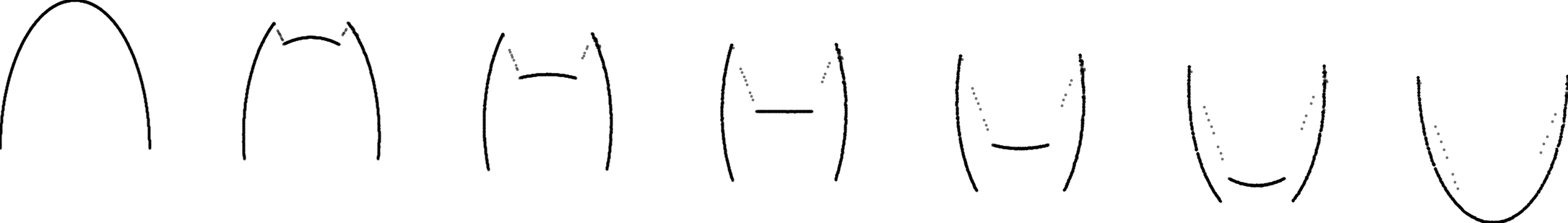}} \\[2mm]
\fbox{\includegraphics[width=0.9\textwidth]{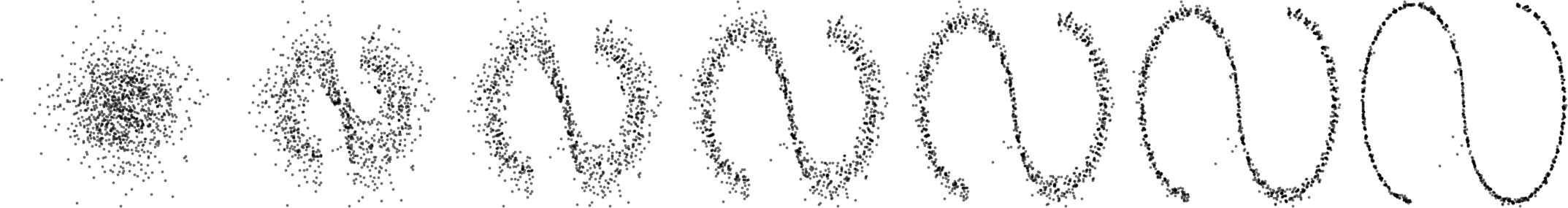}} \\[2mm]
\fbox{\includegraphics[width=0.9\textwidth]{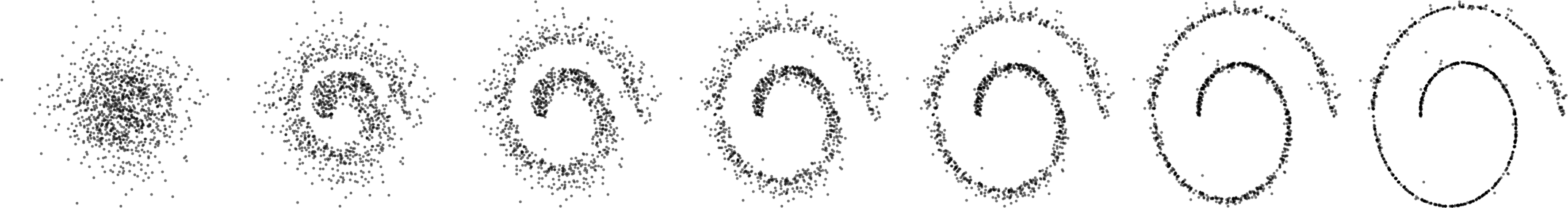}}

\end{document}